\documentclass{article}

     \PassOptionsToPackage{numbers, compress}{natbib}

\usepackage[preprint]{neurips_2021}

\usepackage[utf8]{inputenc} 
\usepackage[T1]{fontenc}    
\usepackage{hyperref}       
\usepackage{url}            
\usepackage{booktabs}       
\usepackage{amsfonts}       
\usepackage{nicefrac}       
\usepackage{microtype}      
\usepackage{xcolor}         
\usepackage{microtype}
\usepackage{graphicx}
\usepackage{subfigure}
\usepackage{booktabs} 
\usepackage{bbm}
\usepackage{mathrsfs}
\usepackage{amsmath}
\usepackage{amssymb}

\usepackage{xcolor}
\usepackage{framed}
\usepackage{wrapfig}
\usepackage{cutwin}

\title{A Hypothesis for the Aesthetic Appreciation \\in Neural Networks}

\author{%
 Xu Cheng\\
 Shanghai Jiao Tong University\\
\texttt{xcheng8@sjtu.edu.cn} \\
 \And
Xin Wang\\ 
Shanghai Jiao Tong University\\
\texttt{xin.wang@sjtu.edu.cn} \\
 \And
Haotian Xue\\ 
Shanghai Jiao Tong University\\
\texttt{xavihart@sjtu.edu.cn} \\
 \And
Zhengyang Liang\\ 
Tongji University\\
\texttt{marxas@tongji.edu.cn} \\
 \And
 Quanshi Zhang\thanks{This research is done under the supervision of Dr. Quanshi Zhang. He is with the John Hopcroft Center and the MoE Key Lab of Artificial Intelligence, AI Institute, at the Shanghai Jiao Tong University, China. Correspondence to: Quanshi Zhang <zqs1022@sjtu.edu.cn>.} \\
 Shanghai Jiao Tong University\\
\texttt{zqs1022@sjtu.edu.cn} \\
}

\begin{document}

\maketitle

\begin{abstract}
 This paper proposes a hypothesis for the aesthetic appreciation that aesthetic images make a neural network strengthen salient concepts and discard inessential concepts.
In order to verify this hypothesis, we use  multi-variate interactions to represent salient concepts and inessential concepts contained in images.
 Furthermore, we design a set of operations to revise images towards more beautiful ones.
 In experiments, we find that the revised images are more aesthetic than the original ones to some extent.
 \emph{The code will be released when the paper is accepted.}
\end{abstract}

\section{Introduction}
The aesthetic appreciation is a high-level and complex issue, which influences choices of human activities, such as the mate selection, art appreciation, and possibly even the moral judgment. From the neurobiological perspective, aesthetic experiences likely emerge from the interaction between the emotion–valuation, sensory–motor, and meaning–knowledge neural systems  \cite{chatterjee2014neuroaesthetics,pearce2016neuroaesthetics}.

Unlike previous studies of aesthetics~\cite{chen2018encoder,chen2018deep,deng2018aesthetic,lu2014rapid,ma2017lamp} learning a model to mimic human annotations of aesthetic levels in a supervised manner, we consider the aesthetic appreciation inherently rooted in the brain.
Specially, we roughly classify the aesthetic appreciation into two types, in terms of whether the aesthetic appreciation has a direct evolutionary utility or not.
Specifically, the aesthetic appreciation with a direct evolutionary utility is mainly concerned with
the survival and reproduction.
For example, the \emph{physical attractiveness}, including the facial beauty,  has a direct evolutionary utility in the partner selection and may reflect the fertility, gene quality, and health \cite{johnston2006mate}.
Whereas, the aesthetic appreciation without a direct evolutionary utility can be regarded as a special case of cognition, which may reduce cognitive costs of humans.
For example, orderly and simple objects may appear to be more beautiful than chaotic or complex objects \cite{birkhoff2013aesthetic}.

Considering the aesthetic appreciation relating to the survival and reproduction is hard to define, in this study, we focus on the second type of the aesthetic appreciation without a direct evolutionary utility.
Specifically, we propose a hypothesis for the aesthetic appreciation that \textbf{aesthetic images make the brain discard inessential ``\emph{concepts}'' and enhance salient ``\emph{concepts}''}, \emph{i.e.} triggering clean and strong signals to the brain.
Note that the concept here is referred to as features for certain shapes or textures, instead of representing semantics.
For example, when we enjoy a piece of melodious music, the Beta wave (a neural oscillation in the brain) may dominate our consciousness, make us concentrate, feel excited, and remain in a state of high cognitive functioning \cite{buzsaki2006rhythms,baumeister2008influence,brookings1996psychophysiological}.
Moreover, based on the hypothesis, we develop a method to revise images towards the more beautiful ones.

However, this hypothesis is difficult to be formulated and to be used to handle images, from the neurobiological perspective. It still remains a challenge to find a convincing and rigorous way of disentangling inessential concepts away from salient concepts encoded by biological neural networks.
Hence, we use an artificial deep neural network (DNN) learned for object classification on a huge dataset (\emph{e.g.} the ImageNet dataset \cite{krizhevsky2012imagenet}) as a substitute for the biological neural network. Given a pre-trained DNN $f$ and an image with $n$ pixels $N=\{1,2,\dots, n\}$, we use a metric in game theory, namely the \emph{multi-variate interaction}, to distinguish
inessential concepts and essential salient concepts encoded in the intermediate layer of the DNN.
Based on the multi-variate interaction,
we further revise the image to ensure that the DNN discards inessential concepts and strengthens essential salient concepts.
Therefore, we can verify the hypothesis for the aesthetic appreciation by checking whether the revised images look more beautiful or not.

To this end, we consider a group of highly interacted pixels as a certain visual concept.
The multi-variate interaction between multiple pixels is defined in game theory, which measures the marginal benefit to the network output when these pixels collaborate \emph{w.r.t.} they work independently.
As a toy example, the nose, mouth, and eyes have a strong interaction in the prediction of the face detection, because they collaborate with each other to make the face a salient concept for inference.

Mathematically, we have proven that the overall utility of the image classification (\emph{i.e.} changes of the classification confidence) can be decomposed into numerical effects of multi-variate interaction concepts $I(S)$, \emph{i.e.}
$\text{\emph{utility}}= \sum_{S\subseteq N,|S|\ge2} I(S)$.
Note that there are $2^{n}-n-1$ concepts theoretically, but only very sparse concepts have significant effects $|I(S)|$ on the network output and a large ratio of concepts have negligible influences $|I(S)|$ on the overall utility.
In this way, salient concepts modeled by the DNN usually refer to concepts with large $|I(S)|$ values, while those with small $|I(S)|$ values are mainly regarded as inessential concepts.

Thus, to verify the proposed hypothesis, we revise images to ensure the DNN strengthens salient concepts and weakens inessential concepts. Experimental results show that the revised images are more aesthetic than the original ones to some extent, which verifies the hypothesis.
Specifically, we propose four independent operations to adjust the hue, saturation, brightness (value), and the sharpness of the image smoothly.
In this way, the image revision will not bring in additional concepts or introduce out-of-distribution information like adversarial examples.

\section{Related work}
The aesthetic appreciation, relating to the enjoyment or study of the beauty, plays a fundamental role in the history and culture of the human being~\cite{bo2018computational}.
The study of aesthetics starting from the philosophy has extended to fields of the psychology \cite{fechner1876vorschule,leder2004model}, neuroscience~\cite{chatterjee2011neuroaesthetics}, and computer
science~\cite{neumann2005defining}.

From the perspective of the \emph{psychology}, previous studies suggested that the aesthetic appreciation could be influenced by the perception, knowledge and content \cite{gracyk2003hume,hammermeister2002german,li2020review,shelley2017concept,vessel2010beauty}.
However, these researches mainly built theoretical models of the aesthetic appreciation and lacked quantitative comparisons between different psychological factors.

\emph{Neuroaesthetics} investigates biological bases of aesthetic experiences when we appraise objects~\cite{brown2011naturalizing}.
\citet{zeki1993vision,zeki1999art} proposed that the function of art was an extension of the visual brain.
\citet{chatterjee2014neuroaesthetics} considered that the aesthetic experience emerged from the interaction among  sensory–motor, emotion–valuation, and meaning–knowledge neural systems.
\citet{brown2011naturalizing} demonstrated that the general reward circuitry of the human produced the pleasure when people looked at beautiful objects.
In conclusion, neuroaesthetics mainly focused on the connection between the activation of specific brain regions and aesthetic experiences
\cite{cela2009sex,epstein1998cortical,greenlee2008functional,jacobsen2017domain,jacobsen2006brain,kirsch2016shaping,luo2019neural}.

\emph{Computational aesthetics} has emerged with the fast development of the digital technology and computer science, which aims to automatically and aesthetically evaluate and generate visual objects like humans. The computational aesthetics can be roughly classified into image aesthetic assessment and image aesthetic manipulation. The task of the image aesthetic assessment is to
distinguish high-quality photos from low-quality photos based on the aesthetic appreciation of human beings.
Previous studies used either hand-crafted features
\cite{bhattacharya2010framework,datta2006studying,dhar2011high,ke2006design,marchesotti2011assessing,marchesotti2013learning,nishiyama2011aesthetic,tong2004classification} or learned deep features \cite{chen2020adaptive,hosu2019effective,kao2017deep,li2020personality,liu2020composition,lu2015rating,mai2016composition,sheng2020revisiting} for the classification of the image quality \cite{lu2014rapid,ma2017lamp}, the aesthetic score regression \cite{kong2016photo}, or the prediction of aesthetic score distributions \cite{jin2016image,wang2017image,wu2011learning}.
Moreover, recent researches also focused on generating captions to describe the aesthetic appreciation of image contents~\cite{chang2017aesthetic,jin2019aesthetic,wang2019neural,zhou2016joint}.
In comparison, the aim of the image aesthetic manipulation is to improve the aesthetic quality of an image.
Previous studies usually enhanced the light, contrast, and composition to make images appear more pleasing \cite{chen2018encoder,chen2018deep,deng2018aesthetic,esmaeili2017fast,kim2020pienet,kim2013optimized,lee2016automatic,lore2017llnet,tu2020image,wei2018good,yan2015change,zeng2019reliable}.
However, these studies were mainly conducted by training a model with annotated aesthetic levels of images, without modeling the aesthetic.
In contrast, we propose a hypothesis for the aesthetic appreciation without human annotations to explain how the DNN encodes images of different aesthetic qualities from the perspective of game theory.
\section{Verification and hypothesis for the aesthetic appreciation}
\label{algorithm}
\textbf{Hypothesis.}
In this paper, we propose a hypothesis for the aesthetic appreciation that \emph{aesthetic images make a neural network strengthen salient concepts and discard inessential concepts.}
In other words,
if an image only contains massive inessential concepts without any dominant salient concepts activating the neural network for inference, then we may consider this image is not so aesthetic.
It is just like the Beta wave (a brainwave) that makes us concentrate on salient concepts, rather than be distracted by the interference of inessential concepts~\cite{dustman1962beta}.

However, a DNN usually uses features to jointly represent different concepts as a mixture model~\cite{zhang2020extracting}.
In order to verify the hypothesis, we must first disentangle feature representations of different concepts.
To this end, we use the multi-variate interaction as a new tool to decompose visual concepts from features encoded in the neural network.
Based on multi-variate interactions, we can further distinguish salient concepts and inessential concepts.
In this way, we find that aesthetic images usually contain more salient concepts than not so aesthetic images, and inessential concepts in the denoised images were usually weakened.
Furthermore, we also revise images by strengthening salient concepts and weakening inessential concepts.
To some degree, the revised images look more aesthetic than the original ones, which verifies the hypothesis.
\subsection{Multi-variate interaction}
\label{multi-variate interaction}
\begin{wrapfigure}{r}{0.35\textwidth}
\centering
\includegraphics[width=0.3\textwidth]{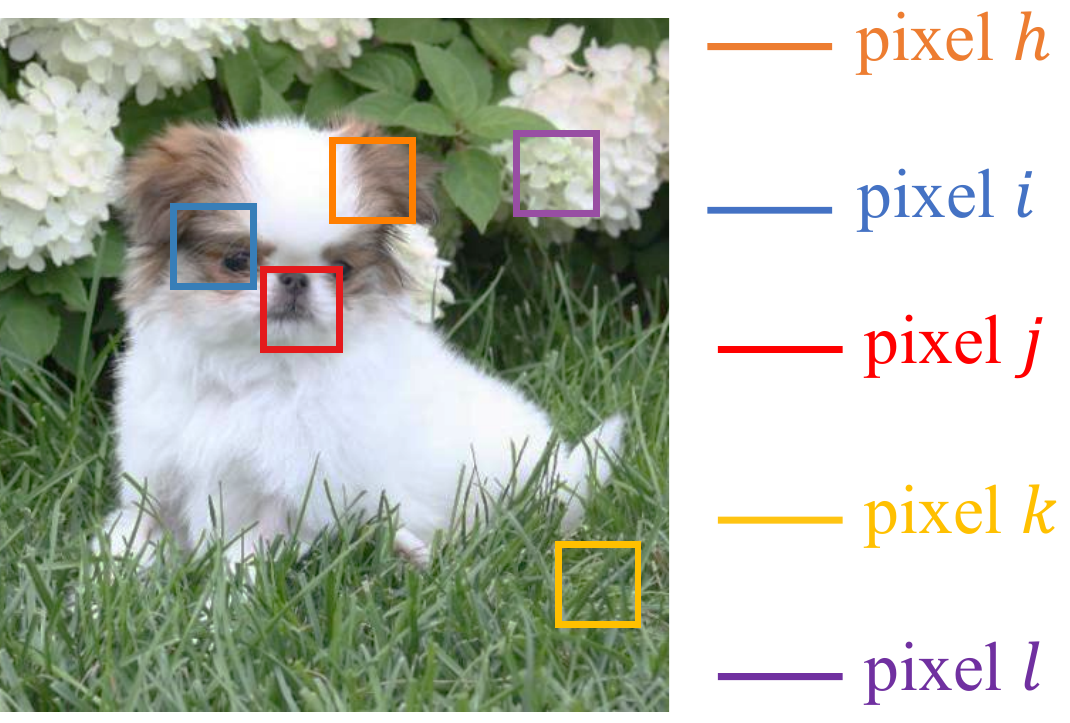}
\vspace{-5pt}
\caption{Visual demonstration for the multi-variate interaction. We use large image regions, instead of pixels, as toy examples to illustrate the basic idea of the multi-variate interaction. In real implementation, we divide the image into $28 \times 28$ grids, and each grid is taken as a ``pixel'' to compute the interaction.}
\label{multi-variate}
\vspace{-20pt}
\end{wrapfigure}

We use the multi-variate interaction between multiple pixels to represent essential salient concepts and inessential concepts of an image in game theory. Given a pre-trained DNN $f$ and the image $x \in  \mathbb{R}^{n}$ of $n$ pixels, $N=\{1,2,\cdots,n\}$, the multi-variate interaction is quantified as the additional utility to the network output when these multiple pixels \emph{collaborate with each other}, in comparison with the case of \emph{they working individually}.
For example, the eyes, a nose, a mouth, and ears of the dog in Fig.~\ref{multi-variate} interact with each other to construct
a meaningful head concept for classification, instead of working independently.
In this way, we can consider a group of strongly interacted pixels as an interaction pattern
(a concept) $S\subseteq N$ for inference, just like the dog head.


\textbf{Definition of the multi-variate interaction (a concept)}.
The interaction inside the pattern $S$ ($|S|\ge2$) is defined in \cite{ren2021learning}.
Let us consider a DNN with a set of pixels, $N=\{h,i,j,k\}$, as a toy example.
These pixels usually collaborate with each other to make impacts on the network output, and a subset of pixels may form a specific interaction pattern (\emph{e.g.} $S=\{h,i,j\} \subseteq N$).
However, if we assume these pixels work individually, then $v(\{i\})=f(\{i\})-f(\emptyset)$ denotes the utility of the pixel $i$ working independently. Here, $f(\{i\})$ represents the output of the DNN only given the information\footnote[1]{We mask pixels to compute $f(S)$. We follow \cite{dabkowski2017real} to set pixels in $N\setminus S$ to baseline values and keep pixels in $S$ unchanged, so as to approximate the removal of pixels in $N\setminus S$.}
of the pixel $i$ by removing the information\textcolor{red}{\footnotemark[1]} of all other pixels in $N\setminus \{i\}$, and $f(\emptyset)$ refers to the output without taking the information\textcolor{red}{\footnotemark[1]} of any pixels into account.
Similarly, $f(S)$ represents the output given the information\textcolor{red}{\footnotemark[1]} of pixels in pattern $S$ and removing the information\textcolor{red}{\footnotemark[1]} of other pixels in $N\setminus S$.
$f(N)$ indicates the output given the information\textcolor{red}{\footnotemark[1]} of all pixels in $N$.
In this way, $U_{\text{individual}}=v(\{h\}) + v(\{i\}) + v(\{j\})$ indicates the utility of all pixels inside the pattern working individually.
On the other hand, the overall utility of the pattern $S$ is given as $U_{\text{joint}}=f(S)-f(\emptyset)$, and obviously $U_{\text{joint}} \ne U_{\text{individual}}$.
Thus, $U_{\text{joint}} - U_{\text{individual}}$ represents the additional utility to the network output caused by collaborations of these pixels. Such additional utility is called the interaction.

More specifically, the overall utility $U_{\text{joint}} - U_{\text{individual}}$ can be decomposed into collaborations between even smaller subsets, such as $\{h, i\}$.
In this way, a visual concept is defined using the elementary interaction between multiple pixels, as follows.
\begin{flalign}
I(S) = \underbrace{f(S) - f(\emptyset)}_{\text{the utility of all pixels in S}} - \sum_{L\subsetneqq S, |L|\ge2} I(L) - \sum_{i\in S} v(\{i\}),
\label{eqn:multi-variate interaction}
\end{flalign}

For example, in Fig.~\ref{multi-variate}, $I(S=\{h,i,j\})$ measures the marginal utility of the interaction within the dog head (utility of the head concept), where utilities of other smaller compositional concepts, \emph{i.e.} $\{\text{eye, nose}\}$, $\{\text{eye, ear}\}$, $\{\text{nose, ear}\}$, $\{\text{eye}\}$, $\{\text{nose}\}$, and $\{\text{ear}\}$, are removed.
Hence, the multi-variate interaction $I(S)$ measures the marginal utility of the pattern $S$ by removing utilities of all smaller patterns, which can be calculated as follows (proved in the supplementary material).
\begin{equation}
I(S) = {\sum}_{L\subseteq S} (-1)^{|S|-|L|} f(L),
\label{eqn:multi-variate interaction cal}
\end{equation}


where $|\cdot|$ denotes the cardinality of the set.

\textbf{Understanding the sparsity of multi-variate interactions, \emph{i.e.} the sparsity of concepts encoded in a DNN.}
Mathematically, we have proven that the classification utilities of all pixels, $f(N)-f(\emptyset)$, can be decomposed into the utility of each interaction pattern (proved in the supplementary material).
In other words, the discrimination power can be decomposed as the sum of classification utilities of massive elementary concepts.
\begin{equation}
\begin{split}
    &f(N) - f(\emptyset)-{\sum}_{i\in N}v(\{i\}) = \sum\nolimits_{\substack{S\subseteq N,\\|S|\geq2}} I(S)
    = \sum\nolimits_{S\in \Omega_{\text{salient}}}I(S) + \sum\nolimits_{S\in \Omega_{\text{inessential}}}I(S).
\end{split}
\label{eqn:utility}
\end{equation}
Therefore, we can divide all concepts (interaction patterns) into a set of salient concepts {\small $\Omega_{\text{salient}}$} that significantly contribute to the network output, and a set of inessential concepts {\small $\Omega_{\text{inessential}}$} with negligible effects on the inference,
{\small $\Omega_{\text{salient}} \cup \Omega_{\text{inessential}}=\{S|S\subseteq N, |S|\ge 2\}$}.
Note that there are $2^{n}-n-1$ concepts, in terms of {\small $\sum\nolimits_{S\subseteq N, |S|\geq2} I(S)$}.
However, we observe that only a few interaction concepts make significant effects $|I(S)|$ on the inference, and these sparse concepts can be regarded as salient concepts.
Most other concepts composed of random pixels make negligible impacts on the network output (\emph{i.e.} $|I(S)| $ is small), and these dense concepts can be considered as inessential concepts, thereby {\small $|\Omega_{\text{inessential}}|\gg|\Omega_{\text{salient}}|$}.

In this way, we consider \textbf{salient concepts usually correspond to interaction patterns with large $|I(S)|$ values, while inessential concepts usually refer to interaction patterns with small $|I(S)|$ values}.
For example, the collaboration between the grass and flower in Fig.~\ref{multi-variate} is not useful for the classification of the dog, \emph{i.e.} being considered as inessential concepts.
In comparison, the dog face is a salient concept with significant influence on the inference.

\subsection{Approximate approach to strengthening salient concepts and weakening inessential concepts}
\label{approximate}
Based on the hypothesis and above analysis, we can improve the aesthetic levels of images by strengthening concepts with large $|I(S)|$ values and weakening concepts with small $|I(S)|$ values, \emph{i.e.} by making concepts more sparse.
However, according to Eq.~\eqref{eqn:utility}, the enumeration of all salient concepts and inessential concepts is an NP-hard problem.
Fortunately, we find a close relationship between interaction patterns (concepts) and Shapley values \cite{Shapley1953}.
Specifically, the term $\Delta f(i,L)$ (defined in Eq.~\eqref{eqn:shapleyvalue}) in the Shapley value can be represented as the sum of some interaction patterns.
In this way, an approximate solution to boost the sparsity of interaction patterns (concepts) is to boost the sparsity of $\Delta f(i,L)$ over pixels $\{i \in N \}$ and contexts $\{L \subseteq N\setminus \{i\}\}$.

\textbf{Definition of Shapley values and $\Delta f(i,L)$}.
Before the optimization of interaction patterns, we first introduce the definition of Shapely values.
The Shapley value is widely considered as an unbiased estimation of the numerical utility \emph{w.r.t.} each pixel in game theory.
Given a trained DNN $f$ and an image $x \in \mathbb{R}^{n}$ with $n$ pixels $N=\{1,2,\cdots,n\}$, some pixels may cooperate to form a context $L\subseteq N$ to influence the output $y= f(x) \in \mathbb{R}$.
To this end, the Shapley value is proposed to fairly divide and assign the overall effects on the network output to each pixel, whose fairness is ensured by \emph{linearity, nullity, symmetry, {\rm and} efficiency} properties~\cite{weber1988probabilistic}.
In this way, the Shapley value, $\phi(i|N)$, represents the utility of the pixel $i$ to the network output as follows.
\begin{equation}
\label{eqn:shapleyvalue}
\phi(i|N)=\sum\nolimits_{L\subseteq N\setminus\{i\}}\frac{(n-|L|-1)!|L|!}{n!}\Big[\Delta f(i,L)\Big],
\end{equation}
where $\Delta f(i,L) =f(L\cup\{i\})-f(L)$ measures the \emph{marginal utility} of the pixel $i$ to the network output, given a set of contextual pixels $L$. Please see the supplementary material for details.

\textbf{Using $\Delta f(i,L)$ to optimize interaction patterns.}
Considering the enumeration of all salient concepts and inessential concepts is NP-hard, it is difficult to directly strengthen salient concepts and weaken inessential concepts.
Fortunately,  we have proven that the term $\Delta f(i,L)$ in the Shapley value can be re-written as the sum of some interaction patterns (proved in the supplementary material).
\begin{equation}
\label{eqn:connection deltaf}
\Delta f(i,L)= v(\{i\}) + {\sum}_{L'\subseteq L, L' \ne \emptyset}  I(S'=L' \cup \{i\}).
\end{equation}
According to empirical observations, $\Delta f(i,L)$ is very sparse among all contexts $L$.
In other words, only a small ratio of contexts $L$ have significant impacts $|\Delta f(i,L)|$ on the network output, and a large ratio of contexts $L$ make negligible impacts $|\Delta f(i,L)|$.
In this way, we can roughly consider salient concepts are usually contained by the term $\Delta f(i,L)$ with a large absolute value, while the term $\Delta f(i,L)$ with a small absolute value, $|\Delta f(i,L)|$, only contains inessential concepts.

Therefore, we define the following loss function to enhance salient concepts contained in $\Delta f(i,L)$ with a large absolute value and weaken inessential concepts included by $\Delta f(i,L)$ with a small absolute value. In this way, we can approximately make interaction patterns (concepts) more sparse.
\begin{equation}
\text{Loss}=
		\mathbb{E}_{r}\bigg[\mathbb{E}_{\substack{ (i_1,L_1)\in \Omega^{\prime}_{\text{inessential}},\\
		|L_1| =r}} \Big[|\Delta f(i_1,L_1)|\Big]
		-\alpha\cdot \mathbb{E}_{\substack{ (i_2,L_2)\in \Omega^{\prime}_{\text{salient}},\\
		|L_2| =r}} \Big[|\Delta f(i_2,L_2)|\Big] \bigg],
\label{eqn:loss}
\end{equation}
where {\small$i_1\in N\setminus L_1$} and {\small$i_2\in N\setminus L_2$}.
{\small$\Omega^{\prime}_{\text{salient}}$} refers to massive coalitions, each representing a set of pixels $(i,L)$ containing salient concepts, thereby generating large values of $|\Delta f(i,L)|$.
Accordingly, {\small$\Omega^{\prime}_{\text{inessential}}$} represents massive coalitions, each representing a set of pixels $(i,L)$ only containing inessential concepts, thereby generating small values of $|\Delta f(i,L)|$.
$\alpha$ is a positive scalar to balance effects between enhancing salient concepts and discarding inessential concepts.

Note that in Eq.~\eqref{eqn:connection deltaf} and Eq.~\eqref{eqn:loss}, we only enumerate contexts $L$ made up of $1\le r \le 0.5n$ pixels. There are two reasons. First, when $r$ is larger, $\Delta f(i,L)$ contains more contexts $L$, and thus it is more difficult to maintain $\Delta f(i,L)$ sparse among all coalitions $\{(i,L)\}$.
This hurts the sparsity assumption for Eq.~\eqref{eqn:connection deltaf}.
Second, collaborations of massive pixels (a large $r$) usually represent interactions between mid-level patterns, instead of representing pixel-wise collaborations, which boost the difficulty of the image revision.
Please see the supplementary material for more discussions.

\textbf{Using high-dimension utility to boost the efficiency of learning interaction patterns.}
The optimization of Eq.~\eqref{eqn:loss} is still inefficient, because
Eq.~\eqref{eqn:loss} only provides a gradient of a scalar output $f(L) \in \mathbb{R}$.
Actually, the theory of Shapley values in Eq.~\eqref{eqn:shapleyvalue} and Eq.~\eqref{eqn:connection deltaf} is compatible with the case of $f(L) \in \mathbb{R}^{d}$ being implemented as a high-dimension feature of the DNN,
thereby receiving gradients of all dimensions of the feature.
In this way, we directly use the intermediate-layer feature $f(L)$ to optimize Eq.~\eqref{eqn:loss}, where the absolute value of the marginal utility $|\Delta f(i,L)|$ is replaced as the $L_1$-norm, $||\Delta f(i,L)||_1$.

\textbf{Determination of salient concepts and inessential concepts.}
 We consider {\small$(i,L) \in \Omega^{\prime}_{\text{inessential}}$} with the smallest ratio $m_1$ of $||\Delta f(i,L)||_{2}$ values as inessential concepts, which are supposed to include inessential interaction patterns {\small$S \in \Omega_{\text{inessential}}$}.
Accordingly, we regard {\small$(i,L) \in \Omega^{\prime}_{\text{salient}}$} with the largest ratio $m_2$ of $||\Delta f(i,L)||_{2}$ values as salient concepts, which are supposed to contain salient interaction patterns {\small$S \in \Omega_{\text{salient}}$}.

\subsection{Operations to revise images}
\label{operation}
In the above section, we propose a loss function to revise images, which ensures the DNN enhances salient concepts and discards inessential concepts.
In this section, we further define a set of operations to revise images to decrease the loss.
\textbf{A key principle for image revision is that the revision should not bring in additional concepts, but just revises existing concepts contained in images.}
Specifically, this principle can be implemented as the following three requirements.\\
1. The image revision is supposed to simply change the hue, saturation, brightness (value), and sharpness of
different regions in the image.\\
2. The revision should ensure the spatial smoothness of images.
In other words, each regional revision of the hue/saturation/brightness/sharpness is supposed to be conducted smoothly over different regions, because dramatic changes of the hue/saturation/brightness/sharpness may bring in new edges
\begin{wrapfigure}{r}{0.5\textwidth}
\centering
\includegraphics[width=0.5\textwidth]{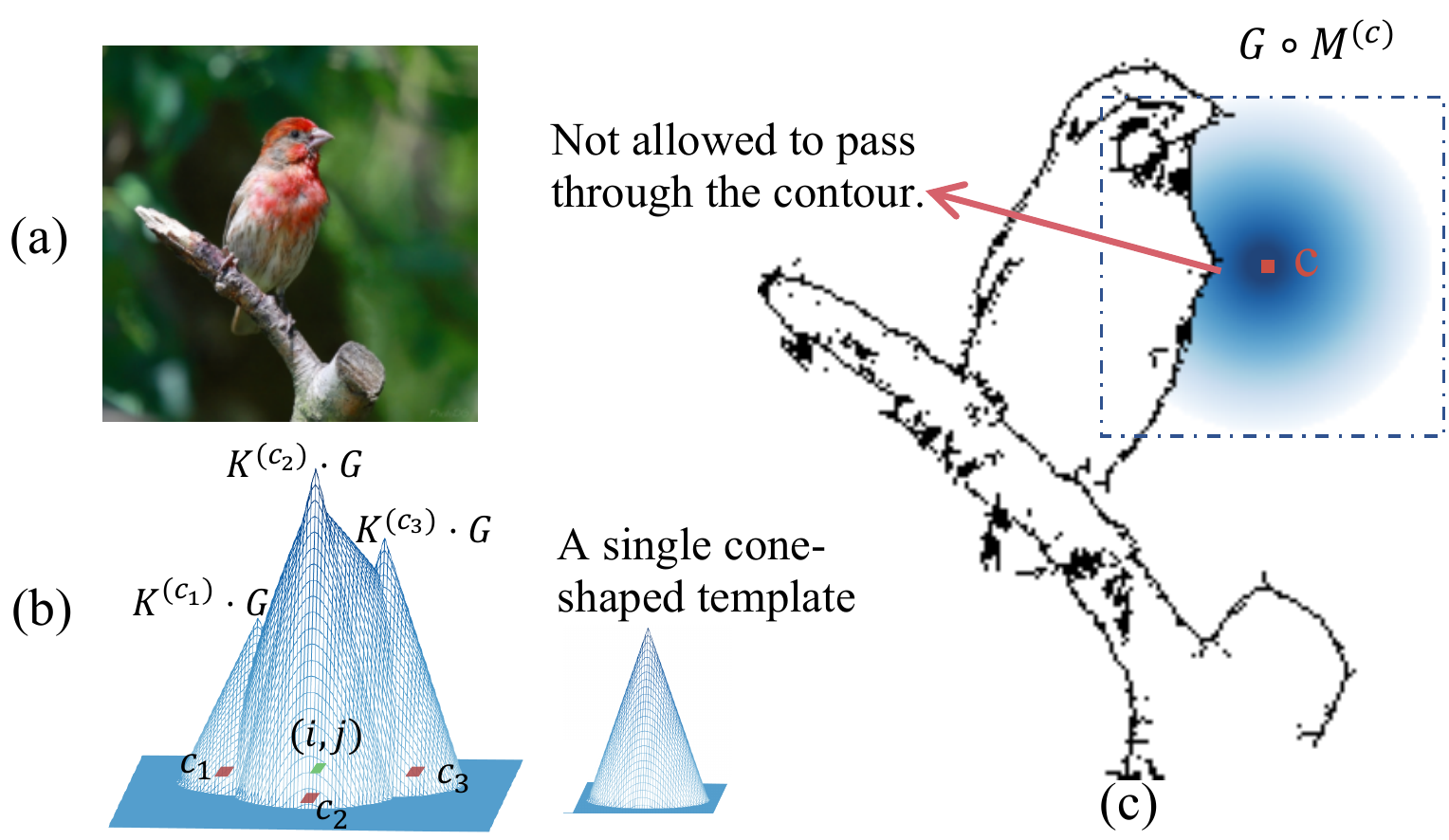}
\vspace{-15pt}
\caption{Visual demonstration for image revision.}
\label{fig:edge}
\vspace{-15pt}
\end{wrapfigure}
between two regions.\\
3. Besides, the revision should also preserve the existing edges.
\emph{I.e.} the image revision is supposed to avoid the color of one object passing through the contour and affecting its neighboring objects.

In this way, we propose four operations to adjust the hue, saturation, brightness, and sharpness of images smoothly,
without damaging existing edges or bringing in new edges.
Thus, the image revision will not generate new concepts or introduce the out-of-distribution features.


\textbf{Operation to adjust the hue}.
Without loss of generality, we first consider the operation to revise the hue of the image $x$ in a small region (with a center $c$).
As Fig.~\ref{fig:edge} (c) shows, we use a cone-shaped template\footnote[2]{We use a cone-shaped template, instead of a Gaussian template, in order to speed up the computation.} $G \in \mathbb{R}^{d'\times d'}$ to smoothly decrease the significance of the revision from the center to the border (requirement 2).
The binary mask $M^{(c)}\in \{0,1\}^{d'\times d'}$ is designed to protect the existing edges (requirement 3).
Specifically, this cone-shaped template $G_{ij} = \max(1-\lambda\cdot ||p_{ij} -c||_{2},0)$ represents the significance of the revision for each pixel in the receptive field, which smooths effects of the image revision.
Here, $||p_{ij} -c||_2$ indicates the Euclidean distance between another pixel $(i,j)$ within the region and the center $c$.
$\lambda$ is a positive scalar to control the range of the receptive field, which is set to $\lambda=1$ for implementation.
Moreover, all pixels blocked by the edge (\emph{e.g.} the object contour) are not allowed to be revised.
$M^{(c)}$ masks image regions that are not allowed to be revised.
As Fig.~\ref{fig:edge}(c) shows, the effects of the sky are not supposed to pass through the contour of the bird and revise the bird head, when the center $c$ locates in the sky.
In this way, the revision of all pixels in the region with the center $c$ is given as
\begin{equation}
\label{eqn:T}
Z^{(c,\text{hue})} =  k^{(c,\text{hue})} \cdot (G \circ M^{(c)}),
\end{equation}
where $\circ$ denotes the element-wise production. We learn $k^{(c,\text{hue})}\in\mathbb R$ via the loss function, in order to
control the significance of the hue revision in this region.

Then, as Fig.~\ref{fig:edge}(b) shows, for each pixel $(i,j)$, the overall effect of its hue revision $\Delta_{ij}^{\text{(hue)}}$ is a mixture of effects from surrounding pixels $N_{ij}$, as follows.
\begin{flalign}
x^{\text{(new, hue)}}_{ij} = \beta^{\text{(hue)}} \cdot \Delta_{ij}^{\text{(hue)}}+x^{\text{(hue)}}_{ij},
\quad
\Delta_{ij}^{\text{(hue)}} =  \tanh({\sum}_{c \in N_{ij}}Z_{ij}^{(c,\text{hue})}), \label{eqn:revise h}
 \end{flalign}
where {\small$ 0\le x^{\text{(hue)}}_{ij} \le 1$} denotes the original hue value of the pixel $(i,j)$ in the image $x$.
$\tanh(\cdot)$ is used to control the value range of the image revision, and the maximum magnitude of hue changes is limited by $ \beta^{\text{(hue)}}$.
Note that the revised hue value {\small$x^{\text{(new, hue)}}_{ij}$} may exceed the range $[0,1]$.
Considering the hue value has a loop effect, the value out of range can be directly modified to $[0,1]$ without being truncated. For example, the hue value of $1.2$ is modified to $0.2$.

\textbf{Operations to adjust the saturation and brightness.}
Without loss of generality, let us take the saturation revision for example.
Considering the value range for saturation is $[0,1]$ without a loop effect, we use the sigmoid function to control the revised saturation value {\small$x^{\text{(new, sat)}}_{ij}$} within the range.
\begin{flalign}
x^{\text{(new, sat)}}_{ij}=  \text{\emph{sigmoid}} \big( \beta^{\text{(sat)}} \cdot \Delta_{ij}^{\text{(sat)}}+ \text{\emph{sigmoid}}^{-1}(x^{\text{(sat)}}_{ij})\big), \quad
\Delta_{ij}^{\text{(sat)}} =  \tanh({\sum}_{c \in N_{ij}}Z_{ij}^{(c,\text{sat})}), \label{eqn:revise s}
\end{flalign}
where {\small $Z^{(c,\text{sat})} =  k^{(c,\text{sat})} \cdot (G \circ M^{(c)})$}, just like Eq.~\eqref{eqn:T}.
Besides, the operation to adjust the brightness is similar to the operation to adjust the saturation.
%
%

\textbf{Operation to sharpen or blur images}.
We propose another operation to sharpen or blur the image in the RGB space.
Just like the hue revision, the sharpening/blurring operation to revise the image can be represented as follows.
 \begin{flalign}
x^{\text{(new, blur/sharp)}}_{ij} =  \Delta_{ij}^{\text{(blur/sharp)}} \cdot \Delta x_{ij} + x_{ij},
\quad
\Delta_{ij}^{\text{(blur/sharp)}} =  \tanh({\sum}_{c \in N_{ij}}Z_{ij}^{(c,\text{blur/sharp})}), \label{eqn:revise sharp}
\end{flalign}
where {\small$\Delta x_{ij} = x^\text{(blur)}_{ij}-x_{ij}$} indicates the pixel-wise change towards blurring the image, where {\small$x^\text{(blur)}$} is obtained by blurring the image using the Gaussian blur operation.
Accordingly, {\small$-\Delta x_{ij}$} describes the pixel-wise change towards sharpening the image.
Fig.~\ref{revised images} and Fig.~\ref{high_low} use heatmaps of {\small$\Delta^{\text{(blur/sharp)}}$}, {\small$\Delta_{ij}^{\text{(blur/sharp)}} \in [-1,1]$}, to represent whether the pixel $(i,j)$ is blurred or sharpened.
If {\small$\Delta_{ij}^{\text{(sharp)}} > 0$}, then this pixel is blurred; otherwise, being sharpened.
{\small $Z^{(c,\text{blur/sharp})} =  k^{(c,\text{blur/sharp})} \cdot (G \circ M^{(c)})$}, just like Eq.~\eqref{eqn:T}.

In this way, we can use the loss function defined in Eq.~\eqref{eqn:loss} to learn the image revision of the hue/saturation/brightness/sharpness to strengthen salient concepts and weaken inessential concepts contained in images, \emph{i.e.}
$\min_{\theta} \text{Loss}$.
Here, {\small$\theta = \{k^{(c,\text{hue})}, k^{(c,\text{sat})}, k^{(c,\text{bright})}, k^{(c,\text{blur/sharp})}|c \in N_\text{center}\}$}, and $N_\text{center}$ denotes a set of central pixels \emph{w.r.t.} different regions in the image $x$.

\section{Experiments}
\label{Experiments}
\subsection{Difference in encoding of concepts between aesthetic images and not so aesthetic images}
\begin{figure}[t]
	\centering
	\includegraphics[width=0.95\linewidth]{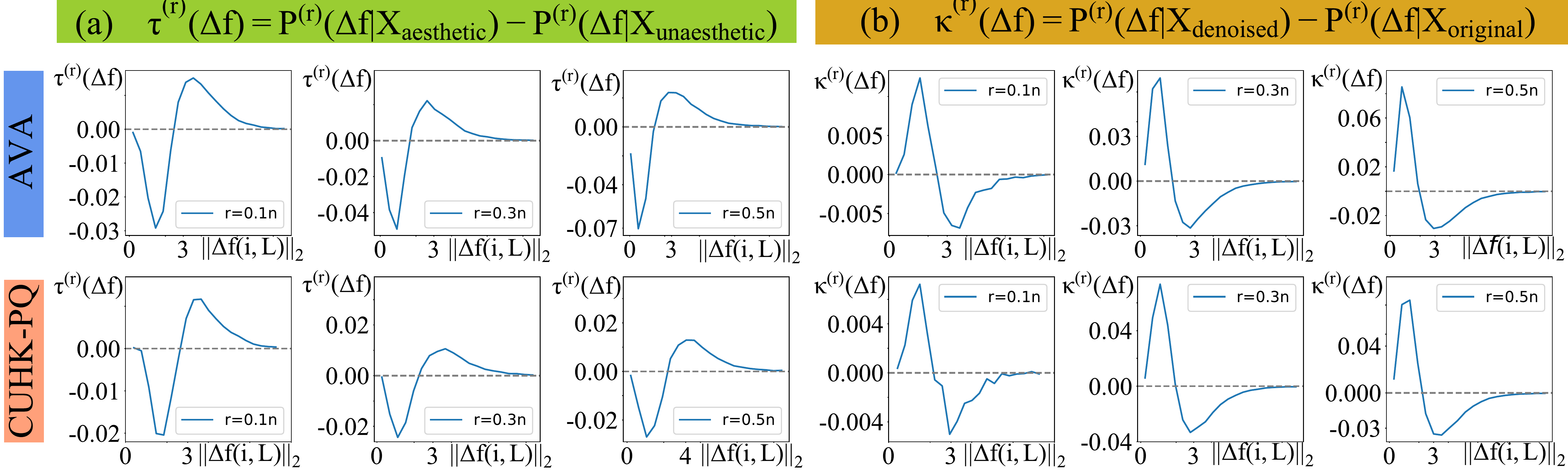}
	\vspace{-8pt}
	\caption{(a) Aesthetic images usually contained more salient concepts than not so aesthetic images.
	 (b) Inessential concepts in the denoised images were weakened.}
	\label{ava}
	\vspace{-12pt}
\end{figure}

To verify the hypothesis, we considered the aesthetic appreciation of images from two aspects, \emph{i.e.}
the beauty of image contents and the noise level.
We considered that images with beautiful contents were usually more aesthetic than images without beautiful contents, and assumed to contain more salient concepts.
On the other hand, we considered
strong noises in images boosted the cognitive burden, and the denoising of an image could increase its aesthetic level.
In this way, we assumed inessential concepts in the denoised images were weakened.

\textbf{Beauty of image contents.}
We used the VGG-16 \cite{simonyan2015very} model trained on the ImageNet dataset \cite{krizhevsky2012imagenet} to evaluate images from the the Aesthetic Visual Analysis (AVA) dataset \cite{murray2012ava} and the CUHK-PhotoQuality (CUHK-PQ) dataset \cite{luo2011content}, respectively.
All aesthetic images and not so aesthetic images had been annotated in these two datasets\footnote[3]{In particular, we considered images in the AVA dataset with the highest aesthetic scores as aesthetic images and regarded images with the lowest aesthetic scores as not so aesthetic images.}.

For verification,
we computed the metric {\small$\tau^{(r)}(\Delta f)=P^{(r)}(\Delta f|X_\text{aesthetic})- P^{(r)}(\Delta f| X_\text{unaesthetic})$}, where {\small$X_\text{aesthetic}$} and {\small$X_\text{unaesthetic}$} referred to a set of  massive aesthetic images and a set of not so aesthetic images, respectively.
{\small$P^{(r)}(\Delta f|X_\text{aesthetic})$} denoted the value distribution of {\small$\Delta f = ||\Delta f(i,L)||_{2}$} among all pixels $\{i\in N\}$ and all contexts {\small$\{L|L\subseteq N\setminus \{i\}, |L|=r\}$} contained by massive aesthetic images
{\small$x \in X_\text{aesthetic}$}.
Similarly, {\small$P^{(r)}(\Delta f|X_\text{unaesthetic})$} represented the value distribution of $\Delta f $ among all pixels $\{i\}$ and all contexts $\{L\}$ contained by not so aesthetic images {\small$x \in X_\text{unaesthetic}$}.
If $\tau^{(r)}(\Delta f) > 0$ for large values of $\Delta f$, it indicated that aesthetic images contained more salient concepts than not so aesthetic images.
In experiments, $f$ was implemented as the feature of the first fully-connected (FC) layer of the VGG-16 model.
Fig.~\ref{ava} (a) shows that aesthetic images usually included more salient concepts than not so aesthetic images.

\textbf{Noises in the image.}
To verify the assumption that inessential concepts in the denoised images were significantly weakened,
we conducted the following experiments.
We used the Gaussian blur operation to smooth images to eliminate noises.
We calculated the metric {\small$\kappa^{(r)}(\Delta f)=P^{(r)}(\Delta f|X_\text{denoised})- P^{(r)}(\Delta f|X_\text{original})$} for verification, where {\small$X_\text{denoised}$} and {\small$X_\text{original}$} referred to a set of the denoised images and a set of the corresponding original images, respectively.
{\small$P^{(r)}(\Delta f|X_\text{denoised})$} represented the value distribution of {\small$\Delta f = ||\Delta f(i,L)||_{2}$} among all pixels $\{i\in N\}$ and all contexts {\small$\{L|L\subseteq N\setminus \{i\}, |L|=r\}$} contained by the denoised images {\small$x \in X_\text{denoised}$}.
Accordingly, {\small$P^{(r)}(\Delta f|X_\text{original})$} denoted the value distribution of $\Delta f$ among all pixels $\{i\}$ and all contexts $\{L\}$ contained by the original images {\small$x \in X_\text{original}$}.
Fig.~\ref{ava} (b) shows inessential concepts in the denoised images were weakened, \emph{i.e.} $\kappa^{(r)}(\Delta f) > 0$ for small values of $\Delta f$.


\begin{figure}[t]
	\centering
	\includegraphics[width=0.90\linewidth]{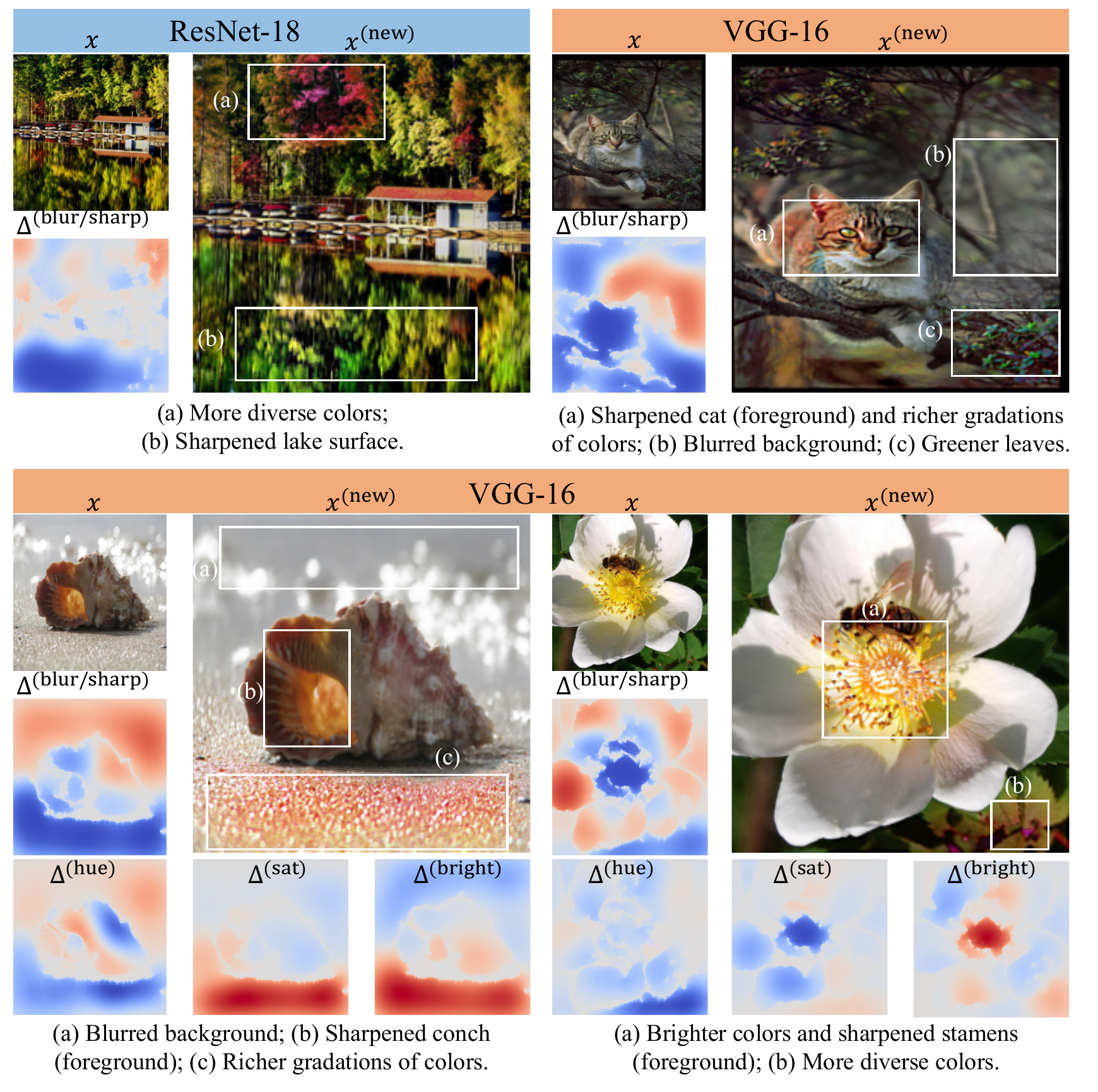}
	\vspace{-8pt}
	\caption{Images revised by using the ResNet-18 and VGG-16 models, respectively. The heatmap {\small$\Delta^{\text{(blur/sharp)}}$} denoted each pixel was blurred or sharpened, where the red color referred to blurring and the blue color corresponded to sharpening.
	Heatmaps {\small$\Delta^{\text{(hue)}}$}, {\small$\Delta^{\text{(sat)}}$}, and {\small$\Delta^{\text{(bright)}}$} indicated the increase (red) and the decrease (blue) of the hue/saturation/brightness revision, respectively. Please see the supplementary material for more results.}
	\label{revised images}
	\vspace{-13pt}
\end{figure}

\subsection{Visualization of image revision}
\label{sec:revised_imgs}
We used the proposed operations to revise images by strengthening salient concepts and weakening inessential concepts, and checked whether the revised images were more beautiful or not.

We computed the loss function based on the VGG-16 and ResNet-18~\cite{he2016deep} models to revise images.
These DNNs were trained on the ImageNet dataset for object classification.
We used the feature after the average pooling operation as $f$ for the ResNet-18 model, and used the feature of the first FC layer as $f$ for the VGG-16 model.
Considering the distribution of {\small$||\Delta f(i,L)||_2$} was long-tailed, we set $m_1 = 60\%$, $m_2 = 15\%$,   and set
 {\small$\beta^{\text{(hue)}}= 0.35, \beta^{\text{(sat)}} = 3$, $\beta^{\text{(bright)}} = 1.5$}.
To balance the effects of strengthening salient concepts and weakening inessential concepts, we set $\alpha = 10$ for ResNet-18 and $\alpha = 8$ for VGG-16, respectively.
We utilized the off-the-shelf edge detection method~\cite{xie2015holistically} to extract the object contour and to calculate the binary mask $M^{(p)}$.
Besides, we used the Gaussian filter with the kernel size $5$ to blur images.

As Fig.~\ref{revised images} shows, the revised images were more aesthetic than the original ones from the following aspects, which verified the hypothesis to some extent. (1) \emph{The revised image usually had richer gradations of the color than the original image}. (2) \emph{The revised image often sharpened the foreground and blurred the background}.
\subsection{Analyzing effects of large-scale concepts and small-scale concepts}
\begin{figure}[t]
	\centering
	\includegraphics[width=\linewidth]{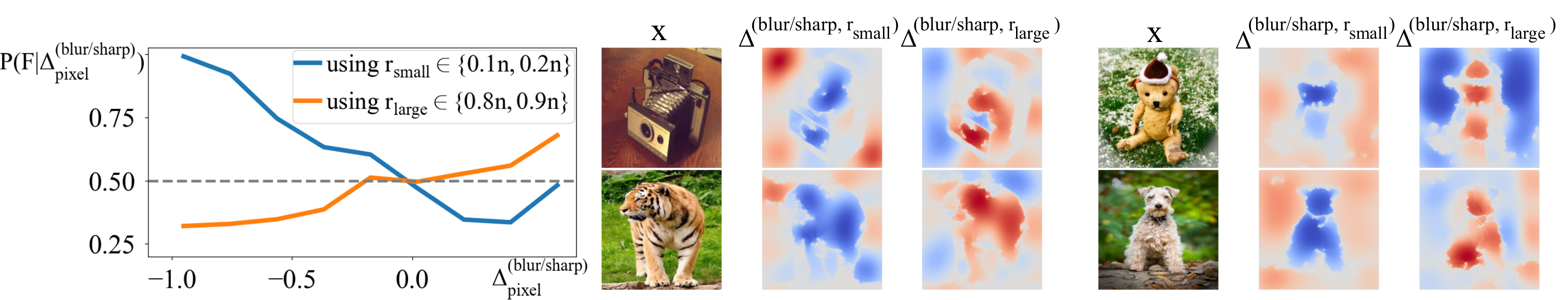}
	\vspace{-15pt}
	\caption{Penalizing small-scale concepts sharpened the foreground and blurred the background of images, while punishing large-scale concepts sharpened the background and blurred the foreground. {\small$\Delta^{(\text{blur/sharp}, r_{\text{small}})}$} and {\small$\Delta^{(\text{blur/sharp}, r_{\text{large}})}$} were calculated by setting {\small$r=r_{\text{small}}$} and {\small$r=r_{\text{large}}$}, respectively.}
	\label{high_low}
	\vspace{-12pt}
\end{figure}

As shown in the supplementary material, the sparsity of {\small$| \Delta f(i,L)|$} could be ensured when we set a small $r$, \emph{i.e.} {\small $1 \le r \le 0.5n$}.
Therefore, we only enumerated contexts $L$ consisting of {\small $1 \le r \le 0.5n$} pixels to represent salient concepts and inessential concepts in Eq.~\eqref{eqn:connection deltaf} and Eq.~\eqref{eqn:loss}.
In this section, we further compared the effects of the image revision between setting a small $r$ value and setting a large $r$ value.
Specifically, we conducted two different experiments.
The first experiment was to penalize small-scale concepts {\small$|| \Delta f(i,L)||_{2}$} of very few contextual pixels {\small$r_{\text{small}} \in \{0.1n, 0.2n\}$}.
The second experiment was to punish large-scale concepts {\small$|| \Delta f(i,L)||_{2}$} of  {\small$r_{\text{large}} \in \{0.8n, 0.9n\}$}.
Fig.~\ref{high_low} shows penalizing small-scale concepts {\small$|| \Delta f(i,L)||_{2}$} usually tended to sharpen the foreground and blurred the background of images.
It was because small-scale concepts usually referred to detailed concepts at the pixel level for the classification of foreground objects, which made the DNN sharpen pixel-wise patterns in the foreground (salient objects) and blur the background (irrelevant objects).
In contrast, punishing large-scale concepts {\small$|| \Delta f(i,L)||_{2}$} were prone to blurring the foreground and sharpening the background.
It was because large-scale concepts usually encoded global contextual knowledge for the inference of objects (\emph{e.g.} the background), which led to sharpening the background.
Although theoretically the background was sometimes also useful for classification, the using of large-scale background knowledge for inference usually caused a higher cognitive burden than using more direct features in the foreground.
Therefore, a common aesthetic appreciation in photography was to blur the background and sharpen the foreground.

\textbf{Implementation details.}
We computed the loss function using the ResNet-18 model trained on the ImageNet dataset, in order to revise images.
As Fig.~\ref{high_low} shows, {\small$\Delta_{\text{pixel}}^{\text{(blur/sharp)}}$} was a pixel in the heatmap {\small$\Delta^{\text{(blur/sharp)}}$} defined in Eq.~\eqref{eqn:revise sharp}, which measured whether this pixel was blurred or sharpened.
Let
{\small$P(\Delta_{\text{pixel}}^{\text{(blur/sharp)}}|F)$} denote the probability distribution of {\small$\Delta_{\text{pixel}}^{\text{(blur/sharp)}}$} among all pixels in the foreground of all images, where {\small$F$} indicated a set of all foreground pixels.
Accordingly, {\small$P(\Delta_{\text{pixel}}^{\text{(blur/sharp)}}|B)$}
represented the probability distribution of {\small$\Delta_{\text{pixel}}^{\text{(blur/sharp)}}$} among all pixels in the background of all images, where {\small$B$} denoted a set of all background pixels.
Here, we used the object bounding box to separate the foreground and the background of each image.
In this way, given a certain value of {\small$\Delta_{\text{pixel}}^{\text{(blur/sharp)}}$}, {\small$ P(F|\Delta_{\text{pixel}}^{\text{(blur/sharp)}})$} described the probability of the pixel with the sharpening/smoothing score {\small$\Delta_{\text{pixel}}^{\text{(blur/sharp)}}$}
locating in the foreground, \emph{i.e.}
{\small$ P(F|\Delta_{\text{pixel}}^{\text{(blur/sharp)}})= P(\Delta_{\text{pixel}}^{\text{(blur/sharp)}}|F) \cdot P(F)/ (P(\Delta_{\text{pixel}}^{\text{(blur/sharp)}}|F) \cdot P(F) +P(\Delta_{\text{pixel}}^{\text{(blur/sharp)}}|B) \cdot P(B))$}.
{\small$P(F)$} and {\small$P(B)$} referred to the prior probability of foreground pixels or background pixels, respectively.
If {\small$ P(F|\Delta_{\text{pixel}}^{\text{(blur/sharp)}}) > 0.5$}, then the pixel with the sharpening/smoothing score {\small$\Delta_{\text{pixel}}^{\text{(blur/sharp)}}$} was more likely to locate in the foreground.
Fig.~\ref{high_low} shows penalizing small-scale concepts {\small$|| \Delta f(i,L)||_{2}$} usually tended to sharpen the foreground and blur the background of images, \emph{i.e.} {\small$P(F|\Delta_{\text{pixel}}^{\text{(blur/sharp)}})>0.5$} when {\small$\Delta_{\text{pixel}}^{\text{(blur/sharp)}} <0$}.
In comparison, punishing large-scale concepts {\small$|| \Delta f(i,L)||_{2}$} often sharpened the background and blurred the foreground.

\section{Conclusion}
\label{conclusion}
In this paper, we propose a hypothesis for the aesthetic appreciation that aesthetic images make a neural network strengthen salient concepts and discard inessential concepts.
In order to verify this hypothesis, we use multi-variate interactions to represent salient concepts and inessential concepts encoded in the DNN.
Furthermore, we develop a new method to revise images, and find that the revised images are more aesthetic than the original ones to some degree.
Nevertheless, in spite of the success in experiments and preliminary verifications, we still need further to verify the trustworthiness of the hypothesis.

\bibliography{aesthetics}

\begin{thebibliography}{77}
\providecommand{\natexlab}[1]{#1}
\providecommand{\url}[1]{\texttt{#1}}
\expandafter\ifx\csname urlstyle\endcsname\relax
  \providecommand{\doi}[1]{doi: #1}\else
  \providecommand{\doi}{doi: \begingroup \urlstyle{rm}\Url}\fi

\bibitem[Baumeister et~al.(2008)Baumeister, Barthel, Geiss, and
  Weiss]{baumeister2008influence}
Jochen Baumeister, T~Barthel, Kurt-Reiner Geiss, and M~Weiss.
\newblock Influence of phosphatidylserine on cognitive performance and cortical
  activity after induced stress.
\newblock \emph{Nutritional neuroscience}, 11\penalty0 (3):\penalty0 103--110,
  2008.

\bibitem[Bhattacharya et~al.(2010)Bhattacharya, Sukthankar, and
  Shah]{bhattacharya2010framework}
Subhabrata Bhattacharya, Rahul Sukthankar, and Mubarak Shah.
\newblock A framework for photo-quality assessment and enhancement based on
  visual aesthetics.
\newblock In \emph{Proceedings of the 18th ACM international conference on
  Multimedia}, pages 271--280, 2010.

\bibitem[Birkhoff(2013)]{birkhoff2013aesthetic}
George~David Birkhoff.
\newblock \emph{Aesthetic measure}.
\newblock Harvard University Press, 2013.

\bibitem[Bo et~al.(2018)Bo, Yu, and Zhang]{bo2018computational}
Yihang Bo, Jinhui Yu, and Kang Zhang.
\newblock Computational aesthetics and applications.
\newblock \emph{Visual Computing for Industry, Biomedicine, and Art},
  1\penalty0 (1):\penalty0 1--19, 2018.

\bibitem[Brookings et~al.(1996)Brookings, Wilson, and
  Swain]{brookings1996psychophysiological}
Jeffrey~B Brookings, Glenn~F Wilson, and Carolyne~R Swain.
\newblock Psychophysiological responses to changes in workload during simulated
  air traffic control.
\newblock \emph{Biological psychology}, 42\penalty0 (3):\penalty0 361--377,
  1996.

\bibitem[Brown et~al.(2011)Brown, Gao, Tisdelle, Eickhoff, and
  Liotti]{brown2011naturalizing}
Steven Brown, Xiaoqing Gao, Loren Tisdelle, Simon~B Eickhoff, and Mario Liotti.
\newblock Naturalizing aesthetics: brain areas for aesthetic appraisal across
  sensory modalities.
\newblock \emph{Neuroimage}, 58\penalty0 (1):\penalty0 250--258, 2011.

\bibitem[Buzsaki(2006)]{buzsaki2006rhythms}
Gyorgy Buzsaki.
\newblock \emph{Rhythms of the Brain}.
\newblock Oxford University Press, 2006.

\bibitem[Cela-Conde et~al.(2009)Cela-Conde, Ayala, Munar, Maest{\'u}, Nadal,
  Cap{\'o}, del R{\'\i}o, L{\'o}pez-Ibor, Ortiz, Mirasso, et~al.]{cela2009sex}
Camilo~J Cela-Conde, Francisco~J Ayala, Enric Munar, Fernando Maest{\'u},
  Marcos Nadal, Miguel~A Cap{\'o}, David del R{\'\i}o, Juan~J L{\'o}pez-Ibor,
  Tom{\'a}s Ortiz, Claudio Mirasso, et~al.
\newblock Sex-related similarities and differences in the neural correlates of
  beauty.
\newblock \emph{Proceedings of the National Academy of Sciences}, 106\penalty0
  (10):\penalty0 3847--3852, 2009.

\bibitem[Chang et~al.(2017)Chang, Lu, and Chen]{chang2017aesthetic}
Kuang-Yu Chang, Kung-Hung Lu, and Chu-Song Chen.
\newblock Aesthetic critiques generation for photos.
\newblock In \emph{Proceedings of the IEEE International Conference on Computer
  Vision}, pages 3514--3523, 2017.

\bibitem[Chatterjee(2011)]{chatterjee2011neuroaesthetics}
Anjan Chatterjee.
\newblock Neuroaesthetics: a coming of age story.
\newblock \emph{Journal of cognitive neuroscience}, 23\penalty0 (1):\penalty0
  53--62, 2011.

\bibitem[Chatterjee and Vartanian(2014)]{chatterjee2014neuroaesthetics}
Anjan Chatterjee and Oshin Vartanian.
\newblock Neuroaesthetics.
\newblock \emph{Trends in cognitive sciences}, 18\penalty0 (7):\penalty0
  370--375, 2014.

\bibitem[Chen et~al.(2018{\natexlab{a}})Chen, Zhu, Papandreou, Schroff, and
  Adam]{chen2018encoder}
Liang-Chieh Chen, Yukun Zhu, George Papandreou, Florian Schroff, and Hartwig
  Adam.
\newblock Encoder-decoder with atrous separable convolution for semantic image
  segmentation.
\newblock In \emph{Proceedings of the European conference on computer vision
  (ECCV)}, pages 801--818, 2018{\natexlab{a}}.

\bibitem[Chen et~al.(2020)Chen, Zhang, Zhou, Lei, Xu, Zheng, and
  Fan]{chen2020adaptive}
Qiuyu Chen, Wei Zhang, Ning Zhou, Peng Lei, Yi~Xu, Yu~Zheng, and Jianping Fan.
\newblock Adaptive fractional dilated convolution network for image aesthetics
  assessment.
\newblock In \emph{Proceedings of the IEEE/CVF Conference on Computer Vision
  and Pattern Recognition}, pages 14114--14123, 2020.

\bibitem[Chen et~al.(2018{\natexlab{b}})Chen, Wang, Kao, and
  Chuang]{chen2018deep}
Yu-Sheng Chen, Yu-Ching Wang, Man-Hsin Kao, and Yung-Yu Chuang.
\newblock Deep photo enhancer: Unpaired learning for image enhancement from
  photographs with gans.
\newblock In \emph{Proceedings of the IEEE Conference on Computer Vision and
  Pattern Recognition}, pages 6306--6314, 2018{\natexlab{b}}.

\bibitem[Dabkowski and Gal(2017)]{dabkowski2017real}
Piotr Dabkowski and Yarin Gal.
\newblock Real time image saliency for black box classifiers.
\newblock \emph{arXiv preprint arXiv:1705.07857}, 2017.

\bibitem[Datta et~al.(2006)Datta, Joshi, Li, and Wang]{datta2006studying}
Ritendra Datta, Dhiraj Joshi, Jia Li, and James~Z Wang.
\newblock Studying aesthetics in photographic images using a computational
  approach.
\newblock In \emph{European conference on computer vision}, pages 288--301.
  Springer, 2006.

\bibitem[Deng et~al.(2018)Deng, Loy, and Tang]{deng2018aesthetic}
Yubin Deng, Chen~Change Loy, and Xiaoou Tang.
\newblock Aesthetic-driven image enhancement by adversarial learning.
\newblock In \emph{Proceedings of the 26th ACM international conference on
  Multimedia}, pages 870--878, 2018.

\bibitem[Dhar et~al.(2011)Dhar, Ordonez, and Berg]{dhar2011high}
Sagnik Dhar, Vicente Ordonez, and Tamara~L Berg.
\newblock High level describable attributes for predicting aesthetics and
  interestingness.
\newblock In \emph{CVPR 2011}, pages 1657--1664. IEEE, 2011.

\bibitem[Dustman et~al.(1962)Dustman, Boswell, and Porter]{dustman1962beta}
Robert~E Dustman, Reed~S Boswell, and Paul~B Porter.
\newblock Beta brain waves as an index of alertness.
\newblock \emph{Science}, 137\penalty0 (3529):\penalty0 533--534, 1962.

\bibitem[Epstein and Kanwisher(1998)]{epstein1998cortical}
Russell Epstein and Nancy Kanwisher.
\newblock A cortical representation of the local visual environment.
\newblock \emph{Nature}, 392\penalty0 (6676):\penalty0 598--601, 1998.

\bibitem[Esmaeili et~al.(2017)Esmaeili, Singh, and Davis]{esmaeili2017fast}
Seyed~A Esmaeili, Bharat Singh, and Larry~S Davis.
\newblock Fast-at: Fast automatic thumbnail generation using deep neural
  networks.
\newblock In \emph{Proceedings of the IEEE Conference on Computer Vision and
  Pattern Recognition}, pages 4622--4630, 2017.

\bibitem[Fechner(1876)]{fechner1876vorschule}
Gustav~Theodor Fechner.
\newblock \emph{Vorschule der aesthetik}, volume~1.
\newblock Breitkopf \& H{\"a}rtel, 1876.

\bibitem[Gracyk(2003)]{gracyk2003hume}
Theodore Gracyk.
\newblock Hume’s aesthetics.
\newblock 2003.

\bibitem[Greenlee and Peter(2008)]{greenlee2008functional}
Mark~W Greenlee and U~Tse Peter.
\newblock Functional neuroanatomy of the human visual system: A review of
  functional mri studies.
\newblock \emph{Pediatric ophthalmology, neuro-ophthalmology, genetics}, pages
  119--138, 2008.

\bibitem[Hammermeister(2002)]{hammermeister2002german}
Kai Hammermeister.
\newblock \emph{The German aesthetic tradition}.
\newblock Cambridge University Press, 2002.

\bibitem[Hosu et~al.(2019)Hosu, Goldlucke, and Saupe]{hosu2019effective}
Vlad Hosu, Bastian Goldlucke, and Dietmar Saupe.
\newblock Effective aesthetics prediction with multi-level spatially pooled
  features.
\newblock In \emph{Proceedings of the IEEE/CVF Conference on Computer Vision
  and Pattern Recognition}, pages 9375--9383, 2019.

\bibitem[Jacobsen and Beudt(2017)]{jacobsen2017domain}
Thomas Jacobsen and Susan Beudt.
\newblock Domain generality and domain specificity in aesthetic appreciation.
\newblock \emph{New Ideas in Psychology}, 47:\penalty0 97--102, 2017.

\bibitem[Jacobsen et~al.(2006)Jacobsen, Schubotz, H{\"o}fel, and
  Cramon]{jacobsen2006brain}
Thomas Jacobsen, Ricarda~I Schubotz, Lea H{\"o}fel, and D~Yves~v Cramon.
\newblock Brain correlates of aesthetic judgment of beauty.
\newblock \emph{Neuroimage}, 29\penalty0 (1):\penalty0 276--285, 2006.

\bibitem[Jin et~al.(2016)Jin, Segovia, and S{\"u}sstrunk]{jin2016image}
Bin Jin, Maria V~Ortiz Segovia, and Sabine S{\"u}sstrunk.
\newblock Image aesthetic predictors based on weighted cnns.
\newblock In \emph{2016 IEEE International Conference on Image Processing
  (ICIP)}, pages 2291--2295. Ieee, 2016.

\bibitem[Jin et~al.(2019)Jin, Wu, Zhao, Li, Zhang, Ge, Zou, Zhou, and
  Zhou]{jin2019aesthetic}
Xin Jin, Le~Wu, Geng Zhao, Xiaodong Li, Xiaokun Zhang, Shiming Ge, Dongqing
  Zou, Bin Zhou, and Xinghui Zhou.
\newblock Aesthetic attributes assessment of images.
\newblock In \emph{Proceedings of the 27th ACM International Conference on
  Multimedia}, pages 311--319, 2019.

\bibitem[Johnston(2006)]{johnston2006mate}
Victor~S Johnston.
\newblock Mate choice decisions: the role of facial beauty.
\newblock \emph{Trends in cognitive sciences}, 10\penalty0 (1):\penalty0 9--13,
  2006.

\bibitem[Kaiming~He and Sun(2016)]{he2016deep}
Shaoqing~Ren Kaiming~He, Xiangyu~Zhang and Jian Sun.
\newblock Deep residual learning for image recognition.
\newblock \emph{In {CVPR}}, 2016.

\bibitem[Kao et~al.(2017)Kao, He, and Huang]{kao2017deep}
Yueying Kao, Ran He, and Kaiqi Huang.
\newblock Deep aesthetic quality assessment with semantic information.
\newblock \emph{IEEE Transactions on Image Processing}, 26\penalty0
  (3):\penalty0 1482--1495, 2017.

\bibitem[Ke et~al.(2006)Ke, Tang, and Jing]{ke2006design}
Yan Ke, Xiaoou Tang, and Feng Jing.
\newblock The design of high-level features for photo quality assessment.
\newblock In \emph{2006 IEEE Computer Society Conference on Computer Vision and
  Pattern Recognition (CVPR'06)}, volume~1, pages 419--426. IEEE, 2006.

\bibitem[Kim et~al.(2020)Kim, Koh, and Kim]{kim2020pienet}
Han-Ul Kim, Young~Jun Koh, and Chang-Su Kim.
\newblock Pienet: Personalized image enhancement network.
\newblock In \emph{European Conference on Computer Vision}, pages 374--390.
  Springer, 2020.

\bibitem[Kim et~al.(2013)Kim, Jang, Sim, and Kim]{kim2013optimized}
Jin-Hwan Kim, Won-Dong Jang, Jae-Young Sim, and Chang-Su Kim.
\newblock Optimized contrast enhancement for real-time image and video
  dehazing.
\newblock \emph{Journal of Visual Communication and Image Representation},
  24\penalty0 (3):\penalty0 410--425, 2013.

\bibitem[Kirsch et~al.(2016)Kirsch, Urgesi, and Cross]{kirsch2016shaping}
Louise~P Kirsch, Cosimo Urgesi, and Emily~S Cross.
\newblock Shaping and reshaping the aesthetic brain: Emerging perspectives on
  the neurobiology of embodied aesthetics.
\newblock \emph{Neuroscience \& Biobehavioral Reviews}, 62:\penalty0 56--68,
  2016.

\bibitem[Kong et~al.(2016)Kong, Shen, Lin, Mech, and Fowlkes]{kong2016photo}
Shu Kong, Xiaohui Shen, Zhe Lin, Radomir Mech, and Charless Fowlkes.
\newblock Photo aesthetics ranking network with attributes and content
  adaptation.
\newblock In \emph{European Conference on Computer Vision}, pages 662--679.
  Springer, 2016.

\bibitem[Krizhevsky et~al.(2012)Krizhevsky, Sutskever, and
  Hinton]{krizhevsky2012imagenet}
Alex Krizhevsky, Ilya Sutskever, and Geoffrey~E Hinton.
\newblock Imagenet classification with deep convolutional neural networks.
\newblock \emph{Advances in neural information processing systems},
  25:\penalty0 1097--1105, 2012.

\bibitem[Leder et~al.(2004)Leder, Belke, Oeberst, and Augustin]{leder2004model}
Helmut Leder, Benno Belke, Andries Oeberst, and Dorothee Augustin.
\newblock A model of aesthetic appreciation and aesthetic judgments.
\newblock \emph{British journal of psychology}, 95\penalty0 (4):\penalty0
  489--508, 2004.

\bibitem[Lee et~al.(2016)Lee, Sunkavalli, Lin, Shen, and
  So~Kweon]{lee2016automatic}
Joon-Young Lee, Kalyan Sunkavalli, Zhe Lin, Xiaohui Shen, and In~So~Kweon.
\newblock Automatic content-aware color and tone stylization.
\newblock In \emph{Proceedings of the IEEE conference on computer vision and
  pattern recognition}, pages 2470--2478, 2016.

\bibitem[Li et~al.(2020)Li, Zhu, Zhao, Ding, and Lin]{li2020personality}
Leida Li, Hancheng Zhu, Sicheng Zhao, Guiguang Ding, and Weisi Lin.
\newblock Personality-assisted multi-task learning for generic and personalized
  image aesthetics assessment.
\newblock \emph{IEEE Transactions on Image Processing}, 29:\penalty0
  3898--3910, 2020.

\bibitem[Li and Zhang(2020)]{li2020review}
Rui Li and Junsong Zhang.
\newblock Review of computational neuroaesthetics: bridging the gap between
  neuroaesthetics and computer science.
\newblock \emph{Brain Informatics}, 7\penalty0 (1):\penalty0 1--17, 2020.

\bibitem[Liu et~al.(2020)Liu, Puri, Kamath, and
  Bhattacharya]{liu2020composition}
Dong Liu, Rohit Puri, Nagendra Kamath, and Subhabrata Bhattacharya.
\newblock Composition-aware image aesthetics assessment.
\newblock In \emph{Proceedings of the IEEE/CVF Winter Conference on
  Applications of Computer Vision}, pages 3569--3578, 2020.

\bibitem[Lore et~al.(2017)Lore, Akintayo, and Sarkar]{lore2017llnet}
Kin~Gwn Lore, Adedotun Akintayo, and Soumik Sarkar.
\newblock Llnet: A deep autoencoder approach to natural low-light image
  enhancement.
\newblock \emph{Pattern Recognition}, 61:\penalty0 650--662, 2017.

\bibitem[Lu et~al.(2014)Lu, Lin, Jin, Yang, and Wang]{lu2014rapid}
Xin Lu, Zhe Lin, Hailin Jin, Jianchao Yang, and James~Z Wang.
\newblock Rapid: Rating pictorial aesthetics using deep learning.
\newblock In \emph{Proceedings of the 22nd ACM international conference on
  Multimedia}, pages 457--466, 2014.

\bibitem[Lu et~al.(2015)Lu, Lin, Jin, Yang, and Wang]{lu2015rating}
Xin Lu, Zhe Lin, Hailin Jin, Jianchao Yang, and James~Z Wang.
\newblock Rating image aesthetics using deep learning.
\newblock \emph{IEEE Transactions on Multimedia}, 17\penalty0 (11):\penalty0
  2021--2034, 2015.

\bibitem[Luo et~al.(2019)Luo, Yu, Li, and Mo]{luo2019neural}
Qiuling Luo, Mengxia Yu, You Li, and Lei Mo.
\newblock The neural correlates of integrated aesthetics between moral and
  facial beauty.
\newblock \emph{Scientific reports}, 9\penalty0 (1):\penalty0 1--10, 2019.

\bibitem[Luo et~al.(2011)Luo, Wang, and Tang]{luo2011content}
Wei Luo, Xiaogang Wang, and Xiaoou Tang.
\newblock Content-based photo quality assessment.
\newblock In \emph{2011 International Conference on Computer Vision}, pages
  2206--2213. IEEE, 2011.

\bibitem[Ma et~al.(2017)Ma, Liu, and Wen~Chen]{ma2017lamp}
Shuang Ma, Jing Liu, and Chang Wen~Chen.
\newblock A-lamp: Adaptive layout-aware multi-patch deep convolutional neural
  network for photo aesthetic assessment.
\newblock In \emph{Proceedings of the IEEE Conference on Computer Vision and
  Pattern Recognition}, pages 4535--4544, 2017.

\bibitem[Mai et~al.(2016)Mai, Jin, and Liu]{mai2016composition}
Long Mai, Hailin Jin, and Feng Liu.
\newblock Composition-preserving deep photo aesthetics assessment.
\newblock In \emph{Proceedings of the IEEE conference on computer vision and
  pattern recognition}, pages 497--506, 2016.

\bibitem[Marchesotti et~al.(2011)Marchesotti, Perronnin, Larlus, and
  Csurka]{marchesotti2011assessing}
Luca Marchesotti, Florent Perronnin, Diane Larlus, and Gabriela Csurka.
\newblock Assessing the aesthetic quality of photographs using generic image
  descriptors.
\newblock In \emph{2011 international conference on computer vision}, pages
  1784--1791. IEEE, 2011.

\bibitem[Marchesotti et~al.(2013)Marchesotti, Perronnin, and
  Meylan]{marchesotti2013learning}
Luca Marchesotti, Florent Perronnin, and France Meylan.
\newblock Learning beautiful (and ugly) attributes.
\newblock In \emph{BMVC}, volume~7, pages 1--11, 2013.

\bibitem[Murray et~al.(2012)Murray, Marchesotti, and Perronnin]{murray2012ava}
Naila Murray, Luca Marchesotti, and Florent Perronnin.
\newblock Ava: A large-scale database for aesthetic visual analysis.
\newblock In \emph{2012 IEEE Conference on Computer Vision and Pattern
  Recognition}, pages 2408--2415. IEEE, 2012.

\bibitem[Neumann et~al.(2005)Neumann, Sbert, Gooch, Purgathofer,
  et~al.]{neumann2005defining}
L~Neumann, M~Sbert, B~Gooch, W~Purgathofer, et~al.
\newblock Defining computational aesthetics.
\newblock \emph{Computational aesthetics in graphics, visualization and
  imaging}, pages 13--18, 2005.

\bibitem[Nishiyama et~al.(2011)Nishiyama, Okabe, Sato, and
  Sato]{nishiyama2011aesthetic}
Masashi Nishiyama, Takahiro Okabe, Imari Sato, and Yoichi Sato.
\newblock Aesthetic quality classification of photographs based on color
  harmony.
\newblock In \emph{CVPR 2011}, pages 33--40. IEEE, 2011.

\bibitem[Pearce et~al.(2016)Pearce, Zaidel, Vartanian, Skov, Leder, Chatterjee,
  and Nadal]{pearce2016neuroaesthetics}
Marcus~T Pearce, Dahlia~W Zaidel, Oshin Vartanian, Martin Skov, Helmut Leder,
  Anjan Chatterjee, and Marcos Nadal.
\newblock Neuroaesthetics: The cognitive neuroscience of aesthetic experience.
\newblock \emph{Perspectives on psychological science}, 11\penalty0
  (2):\penalty0 265--279, 2016.

\bibitem[Ren et~al.(2021)Ren, Zhou, Chen, and Zhang]{ren2021learning}
Jie Ren, Zhanpeng Zhou, Qirui Chen, and Quanshi Zhang.
\newblock Learning baseline values for shapley values, 2021.

\bibitem[Shapley(1953)]{Shapley1953}
L.~Shapley.
\newblock A value for n-person games.
\newblock 1953.

\bibitem[Shelley(2017)]{shelley2017concept}
James Shelley.
\newblock The concept of the aesthetic.
\newblock 2017.

\bibitem[Sheng et~al.(2020)Sheng, Dong, Chai, Wang, Zhou, Huang, Hu, Ji, and
  Ma]{sheng2020revisiting}
Kekai Sheng, Weiming Dong, Menglei Chai, Guohui Wang, Peng Zhou, Feiyue Huang,
  Bao-Gang Hu, Rongrong Ji, and Chongyang Ma.
\newblock Revisiting image aesthetic assessment via self-supervised feature
  learning.
\newblock In \emph{Proceedings of the AAAI Conference on Artificial
  Intelligence}, volume~34, pages 5709--5716, 2020.

\bibitem[Simonyan and Zisserman(2015)]{simonyan2015very}
Karen Simonyan and Andrew Zisserman.
\newblock Very deep convolutional networks for large-scale image recognition.
\newblock \emph{In {ICLR}}, 2015.

\bibitem[Tong et~al.(2004)Tong, Li, Zhang, He, and
  Zhang]{tong2004classification}
Hanghang Tong, Mingjing Li, Hong-Jiang Zhang, Jingrui He, and Changshui Zhang.
\newblock Classification of digital photos taken by photographers or home
  users.
\newblock In \emph{Pacific-Rim Conference on Multimedia}, pages 198--205.
  Springer, 2004.

\bibitem[Tu et~al.(2020)Tu, Niu, Zhao, Cheng, and Zhang]{tu2020image}
Yi~Tu, Li~Niu, Weijie Zhao, Dawei Cheng, and Liqing Zhang.
\newblock Image cropping with composition and saliency aware aesthetic score
  map.
\newblock In \emph{Proceedings of the AAAI Conference on Artificial
  Intelligence}, volume~34, pages 12104--12111, 2020.

\bibitem[Vessel and Rubin(2010)]{vessel2010beauty}
Edward~A Vessel and Nava Rubin.
\newblock Beauty and the beholder: Highly individual taste for abstract, but
  not real-world images.
\newblock \emph{Journal of vision}, 10\penalty0 (2):\penalty0 18--18, 2010.

\bibitem[Wang et~al.(2019)Wang, Yang, Zhang, and Zhang]{wang2019neural}
Wenshan Wang, Su~Yang, Weishan Zhang, and Jiulong Zhang.
\newblock Neural aesthetic image reviewer.
\newblock \emph{IET Computer Vision}, 13\penalty0 (8):\penalty0 749--758, 2019.

\bibitem[Wang et~al.(2017)Wang, Liu, Chang, Dolcos, Beck, and
  Huang]{wang2017image}
Zhangyang Wang, Ding Liu, Shiyu Chang, Florin Dolcos, Diane Beck, and Thomas
  Huang.
\newblock Image aesthetics assessment using deep chatterjee's machine.
\newblock In \emph{2017 International Joint Conference on Neural Networks
  (IJCNN)}, pages 941--948. IEEE, 2017.

\bibitem[Weber(1988)]{weber1988probabilistic}
Robert~J Weber.
\newblock Probabilistic values for games, the shapley value. essays in honor of
  lloyd s. shapley (ae roth, ed.), 1988.

\bibitem[Wei et~al.(2018)Wei, Zhang, Shen, Lin, Mech, Hoai, and
  Samaras]{wei2018good}
Zijun Wei, Jianming Zhang, Xiaohui Shen, Zhe Lin, Radom{\'\i}r Mech, Minh Hoai,
  and Dimitris Samaras.
\newblock Good view hunting: Learning photo composition from dense view pairs.
\newblock In \emph{Proceedings of the IEEE Conference on Computer Vision and
  Pattern Recognition}, pages 5437--5446, 2018.

\bibitem[Wu et~al.(2011)Wu, Hu, and Gao]{wu2011learning}
Ou~Wu, Weiming Hu, and Jun Gao.
\newblock Learning to predict the perceived visual quality of photos.
\newblock In \emph{2011 International Conference on Computer Vision}, pages
  225--232. IEEE, 2011.

\bibitem[Xie and Tu(2015)]{xie2015holistically}
Saining Xie and Zhuowen Tu.
\newblock Holistically-nested edge detection.
\newblock In \emph{Proceedings of the IEEE international conference on computer
  vision}, pages 1395--1403, 2015.

\bibitem[Yan et~al.(2015)Yan, Lin, Kang, and Tang]{yan2015change}
Jianzhou Yan, Stephen Lin, Sing~Bing Kang, and Xiaoou Tang.
\newblock Change-based image cropping with exclusion and compositional
  features.
\newblock \emph{International Journal of Computer Vision}, 114\penalty0
  (1):\penalty0 74--87, 2015.

\bibitem[Zeki(1993)]{zeki1993vision}
Semir Zeki.
\newblock \emph{A vision of the brain.}
\newblock Blackwell scientific publications, 1993.

\bibitem[Zeki(1999)]{zeki1999art}
Semir Zeki.
\newblock Art and the brain.
\newblock \emph{Journal of Consciousness Studies}, 6\penalty0 (6-7):\penalty0
  76--96, 1999.

\bibitem[Zeng et~al.(2019)Zeng, Li, Cao, and Zhang]{zeng2019reliable}
Hui Zeng, Lida Li, Zisheng Cao, and Lei Zhang.
\newblock Reliable and efficient image cropping: A grid anchor based approach.
\newblock In \emph{Proceedings of the IEEE/CVF Conference on Computer Vision
  and Pattern Recognition}, pages 5949--5957, 2019.

\bibitem[Zhang et~al.(2020)Zhang, Wang, Cao, Wu, Shi, and
  Zhu]{zhang2020extracting}
Quanshi Zhang, Xin Wang, Ruiming Cao, Ying~Nian Wu, Feng Shi, and Song-Chun
  Zhu.
\newblock Extracting an explanatory graph to interpret a cnn.
\newblock \emph{IEEE transactions on pattern analysis and machine
  intelligence}, 2020.

\bibitem[Zhou et~al.(2016)Zhou, Lu, Zhang, and Wang]{zhou2016joint}
Ye~Zhou, Xin Lu, Junping Zhang, and James~Z Wang.
\newblock Joint image and text representation for aesthetics analysis.
\newblock In \emph{Proceedings of the 24th ACM international conference on
  Multimedia}, pages 262--266, 2016.

\end{thebibliography}
\bibliographystyle{plainnat}

\newpage
\appendix
\section{Proof of the closed-form solution to the multi-variate interaction in Eq. (2)}
The multi-variate interaction $I(S)$ defined in \cite{ren2021learning} measures the marginal utility from the collaboration of input pixels in the pattern $S$, in comparison with the utility when these pixels work individually.
Specifically, let $f(S)-f(\emptyset)$ denote the overall utility received from all pixels in $S$. Then, we remove utilities of individual pixels without collaborations, \emph{i.e.} $\sum_{i\in S} v(\{i\})=\sum_{i\in S} (f(\{i\})-f(\emptyset))$, and further remove the marginal utilities owing to collaborations of all smaller compositional patterns $L$ within $S$, \emph{i.e.} $\big\{L\subsetneq S|\vert L\vert\ge 2\big\}$. 
In this way, the multi-variate interaction $I(S)$ defined in \cite{ren2021learning} is given as follows.
\begin{equation}
I(S) = \underbrace{f(S) - f(\emptyset)}_{\text{the utility of all pixels in S}} - \sum_{L\subsetneqq S, |L|\ge2} I(L) - \sum_{i\in S} v(\{i\}).
\label{eqn:define multi-variate interaction}
\end{equation}
The closed-form solution for Eq.~\eqref{eqn:define multi-variate interaction} is proven as follows.
\begin{equation}
I(S) = {\sum}_{L\subseteq S} (-1)^{|S|-|L|} f(L).
\label{eqn:multi-variate interaction cal}
\end{equation}
$\bullet\;$\emph{Proof}: 
Here, let $s=|S|$, $l=|L|$, and $s'=|S'|$ for simplicity. If $s=2$, then
\begin{align*}
	I(S) &= f(S)-f(\emptyset)-\sum_{L\subsetneqq S,l\ge 2} I(L) -\sum_{i\in S} v(\{i\})\\
	&= f(S)-f(\emptyset)-\sum_{i\in S}v(\{i\})\\
	&= f(S)-f(\emptyset)-\sum_{i\in S}\left[f(\{i\})-f(\emptyset)\right]\\
	&= f(S)-\sum_{i\in S} f(\{i\}) +f(\emptyset)\\
	&= \sum_{L\subseteq S}(-1)^{s-l} f(L)	
\end{align*}

Let us assume that if $2\le s' <s$, $I(S^{\prime})=\sum_{L\subseteq S^{\prime}} (-1)^{s^{\prime}-l}f(L)$.
Then, we use the mathematical induction to prove $I(S)=\sum_{L\subseteq S} (-1)^{s-l}f(L)$, where $ S^{\prime} \subsetneqq S$.
\begin{small}
\begin{align*}
	I(S=S^{\prime}\cup\{i\}) =& f(S^{\prime}\cup \{i\}) - f(\emptyset) -\sum_{\substack{L\subsetneqq S^{\prime}\cup\{i\},\\l\ge 2}} I(L) - \sum_{j\in S^{\prime}\cup\{i\}} v(\{j\})\qquad \%\;\text{Let }S=S^{\prime} \cup\{i\}\\
	=& f(S^{\prime}\cup \{i\}) - f(\emptyset) -\sum_{\substack{L\subsetneqq S^{\prime}\cup\{i\},\\l\ge 2}} \sum_{K\subseteq L} (-1)^{l-k} f(K)- \sum_{j\in S^{\prime}\cup\{i\}} v(\{j\})\\
	=& f(S^{\prime}\cup \{i\}) - f(\emptyset) -\sum_{K\subsetneqq S^{\prime}}\;\sum_{\substack{T\subseteq S^{\prime}\cup\{i\}\setminus K,\\t+s^{\prime}\ge 2}} (-1)^{t} f(K)\\
	&- \sum_{j\in S^{\prime}\cup\{i\}} v(\{j\}) \qquad \%\;\text{Let } T=L\setminus K\\
	=& f(S^{\prime}\cup \{i\}) - f(\emptyset) -\bigg[\sum_{t=2}^{s^{\prime}} C^{t}_{s^{\prime}+1} (-1)^{t} f(\emptyset) + \sum_{j\in S^{\prime}\cup\{i\}}\sum_{t=1}^{s^{\prime}-1} C^{t}_{s^{\prime}} (-1)^{t} f(\{j\}) \\
	&+ \sum_{\substack{K\subsetneqq S^{\prime}\cup\{i\},\\k\ge 2}} \sum_{t=0}^{s^{\prime}-k} C^{t}_{s^{\prime}+1-k} (-1)^{t} f(K)\bigg] - \sum_{j\in S^{\prime}\cup\{i\}} v(\{j\}) \\
	=& f(S^{\prime}\cup \{i\}) - f(\emptyset) - (s^{\prime}-(-1)^{s^{\prime}+1})v(\emptyset) + \sum_{j\in S\cup\{i\}} (1+(-1)^{s^{\prime}}) f(\{j\})\\
	&+ \sum_{\substack{K\subsetneqq S^{\prime}\cup\{i\},\\k\ge 2}}(-1)^{s^{\prime}+1-k} f(K) - \sum_{j\in S^{\prime}\cup\{i\}} \left[f(\{i\})-f(\emptyset)\right]\\
	=& f(S^{\prime}\cup\{i\}) +\sum_{j\in S^{\prime}\cup\{i\}} (-1)^{s^{\prime}} f(\{j\}) + (-1)^{s^{\prime}+1} f(\emptyset) + \sum_{\substack{K\subsetneqq S^{\prime}\cup\{i\},\\k\ge 2}}(-1)^{s^{\prime}+1-k} f(K) \\
	=& \sum_{K\subseteq S^{\prime}\cup\{i\}} (-1)^{s^{\prime}+1-k} f(K)
	=  \sum_{K\subseteq S} (-1)^{s-k} f(K)
\end{align*}
\end{small}
In this way, Equation~\eqref{eqn:multi-variate interaction cal} is proven as the closed-form solution to the definition of the multi-variate interaction $I(S)$ in Equation~\eqref{eqn:define multi-variate interaction}.

\section{Proof of the decomposition of classification utilities in Eq. (3)}
Mathematically, we have proven that the classification utilities of all pixels, $f(N)-f(\emptyset)$, can be decomposed as the sum of utilities of different interaction patterns.
In other words, the discrimination power can be decomposed as the sum of classification utilities of massive elementary concepts.
\begin{equation}
\begin{split}
    f(N) - f(\emptyset)-{\sum}_{i\in N}v(\{i\}) = \sum\nolimits_{\substack{S\subseteq N,\\|S|\geq2}} I(S)
\end{split}
\label{eqn:utility}
\end{equation}
$\bullet\;$\emph{Proof}: 
\begin{align*}
	\sum_{\substack{S\subseteq N,\\|S|\geq2}} I(S)+f(\emptyset)+\sum_{i\in N}v(\{i\})
	&= \sum_{\substack{S\subseteq N,\\|S|\geq2}}\sum_{L\subseteq S} (-1)^{|S|-|L|} f(L)+f(\emptyset)+\sum_{i\in N}\left[f(\{i\})-f(\emptyset)\right] \\
	&= \sum_{S\subseteq N}\sum_{L\subseteq S} (-1)^{|S|-|L|} f(L) \\
	&= \sum_{L\subseteq N} \sum_{K\subseteq N\setminus L} (-1)^{|K|} f(L) \qquad \%\;\text{Let }K=S\setminus L\\
	&= \sum_{L\subseteq N}\left[\sum_{|K|=0}^{n-|L|}C^{|K|}_{n-|L|} (-1)^{|K|}\right] f(L)
	= f(N) 
\end{align*}

\section{Detailed introduction of Shapley values}
The Shapley value~\cite{Shapley1953} defined in game theory is widely considered as an unbiased estimation of the numerical utility \emph{w.r.t.} each input pixel. 
Given a trained DNN $f$ and an image $x \in \mathbb{R}^{n}$ with $n$ pixels $N=\{1,2,\cdots,n\}$, some pixels may cooperate to form a context $L\subseteq N$ to influence the output $y= f(x) \in \mathbb{R}$.
To this end, the Shapley value is proposed to fairly divide and assign the overall effects on the network output to each pixel.
In this way, the Shapley value, $\phi(i|N)$, represents the utility of the pixel $i$ to the network output, as follows.
\begin{equation}
\label{eqn:shapleyvalue}
\phi(i|N)=\sum\nolimits_{L\subseteq N\setminus\{i\}}\frac{(n-|L|-1)!|L|!}{n!}\Big[\Delta f(i,L)\Big],
\end{equation}
where $\Delta f(i,L) \overset{\text{def}}{=} f(L\cup\{i\})-f(L)$ measures the marginal utility of the pixel $i$ to the network output, given a set of contextual pixels $L$. 

\citet{weber1988probabilistic} has proven that the Shapley value is a unique method to fairly allocate the overall utility to each pixel that satisfies following properties. 

\textbf{(1) Linearity property}: If two independent DNNs can be merged into one DNN, then the Shapley value of the new DNN also can be merged, \emph{i.e.} $\forall i \in N$, $\phi_{u}(i|N)=\phi_{v}(i|N)+\phi_{w}(i|N)$; $\forall c \in \mathbb{R}$, $\phi_{c \cdot u} (i|N)= c\cdot \phi_{u}(i|N)$.

\textbf{(2) Nullity property}: The dummy pixel $i$ is defined as a pixel satisfying $\forall S\subseteq N\setminus\{i\}$, $v(S\cup\{i\})=v(S)+v(\{i\})$, which indicates that the pixel $i$ has no interactions with other pixels in $N$, $\phi(i|N)=v(\{i\})- v(\emptyset)$.

\textbf{(3) Symmetry property}: If $\forall S\subseteq N\setminus\{i,j\}$, $v(S\cup\{i\})=v(S\cup\{j\})$, then $\phi(i|N)=\phi(j|N)$.

\textbf{(4) Efficiency property}: Overall utility can be assigned to all pixels, $\sum_{i\in N}\phi(i|N)=v(N) - v(\emptyset)$.

\section{Proof of Eq. (5)}
We have proven that the term $\Delta f(i,L)$ in the Shapley value can be re-written as the sum of some interaction patterns.
\begin{equation}
\label{eqn:connection deltaf}
\Delta f(i,L)= v(\{i\}) + {\sum}_{L^{'}\subseteq L, L^{'} \ne \emptyset}  I(S'=L' \cup \{i\}).
\end{equation}
$\bullet\;$\emph{Proof}: 
\begin{small}
\begin{align*}
\text{right} &= v(\{i\}) + {\sum}_{L^{'}\subseteq L, L^{'}\ne \emptyset}  I(S'=L' \cup \{i\})\\
	&= v(\{i\}) +\sum_{\substack{L^{'}\subseteq L, \\L^{'} \ne \emptyset}}\Big[\sum_{K\subseteq L^{'}}(-1)^{|L'|+1-|K|} f(K) \\
	&\quad+ \sum_{K\subseteq L^{'}}(-1)^{|L'|-|K|} f(K\cup\{i\}) \Big] \quad\%\;\text{Based on Eq.~\eqref{eqn:multi-variate interaction cal}}\\
	&=v(\{i\}) + \sum_{L^{'}\subseteq L, L^{'} \ne \emptyset}\,\sum_{K\subseteq L^{'}} (-1)^{|L'|-|K|}\left[f(K\cup\{i\}) - f(K)\right] \\
	&= \sum_{L^{'}\subseteq L}\,\sum_{K\subseteq L^{'}} (-1)^{|L'|-|K|}\Delta f(i,K)\\
	&= \sum_{K\subseteq L} \sum_{P\subseteq L\setminus K} (-1)^{|P|} \Delta f(i,K) \qquad \quad\%\;\text{Let } P=L'\setminus K\\
	&= \sum_{K\subseteq L} \left(\sum_{p=0}^{|L|-|K|} (-1)^p C^p_{|L|-|K|}\right)  \Delta f(i,K) \qquad \quad\%\;\text{Let } p=|P|\\
	&= \sum_{K\subsetneqq L} 0\cdot \Delta f(i,K) + \sum_{K= L} \left(\sum_{p=0}^{|L|-|K|} (-1)^p C^p_{|L|-|K|}\right) \Delta f(i,K) \\
	&= \Delta f(i,L) = \text{left}
\end{align*}
\end{small}

\section{Discussions on the enumeration of contexts}
In this section, we analyze reasons for only enumerating contexts $L$ consisting of $1\le r \le 0.5n$ pixels in Eq. (5) and Eq. (6).
In summary, there are two reasons.

\textbf{First, a large $r$ hurts the sparsity assumption.} In this paper, we assume that the marginal utility $\Delta f(i,L)$ is very sparse among all contexts $\{L\}$. In other words, only a small ratio of contexts $\{L\}$ have significant impacts $|\Delta f(i,L)|$ on the network output, and most contexts $\{L\}$ make negligible impacts $|\Delta f(i,L)|$.
If $r$ is larger, according to Eq.~\eqref{eqn:connection deltaf}, $\Delta f(i,L)$ contains more contexts $\{L\}$ .
For example, if $r=n$, there are $2^{n} -n -1$ contexts $\{L\}$ in total.
It is more difficult to maintain $\Delta f(i,L)$ sparse among these massive coalitions $\{(i,L)\}$.
It is because, based on Eq.~\eqref{eqn:connection deltaf}, $\Delta f(i,L)$ is represented as the sum of massive interaction patterns, and then we cannot guarantee only a small ratio of contexts $\{L\}$ have significant impacts $|\Delta f(i,L)|$ on the network output.
In this way, a large $r$ hurts the sparsity assumption.

\textbf{Second, a large $r$ boots the difficulty of the image revision.} 
It is because contexts $L$ consisting of massive pixels (a large $r$) are usually encoded as interactions between middle-scale concepts or between large-scale concepts (\emph{e.g.} background), rather than pixel-level collaborations. In this way, it is difficult to revise images at the pixel level by 
penalizing such middle/large-scale concepts.
Whereas, contexts consisting of a few pixels are mainly encoded as interactions between small-scale concepts at the pixel level, such a local texture or a local object.
Thus, punishing small-scale concepts is more likely to revise the image in detail at the pixel level than punishing large-scale concepts.
Moreover, experiments in Section 4.3 of the main paper also indicated that setting a large $r$ could boost the difficulty of the image revision. 
Punishing large-scale concepts often sharpened the background and blurred the foreground.
It is because large-scale concepts usually encoded global background knowledge as a supplement to the classification, instead of encoding the detailed foreground knowledge for classification.
\section{More visualization results of image revision}
\begin{figure}[h!]
	\centering
	\includegraphics[width=\linewidth]{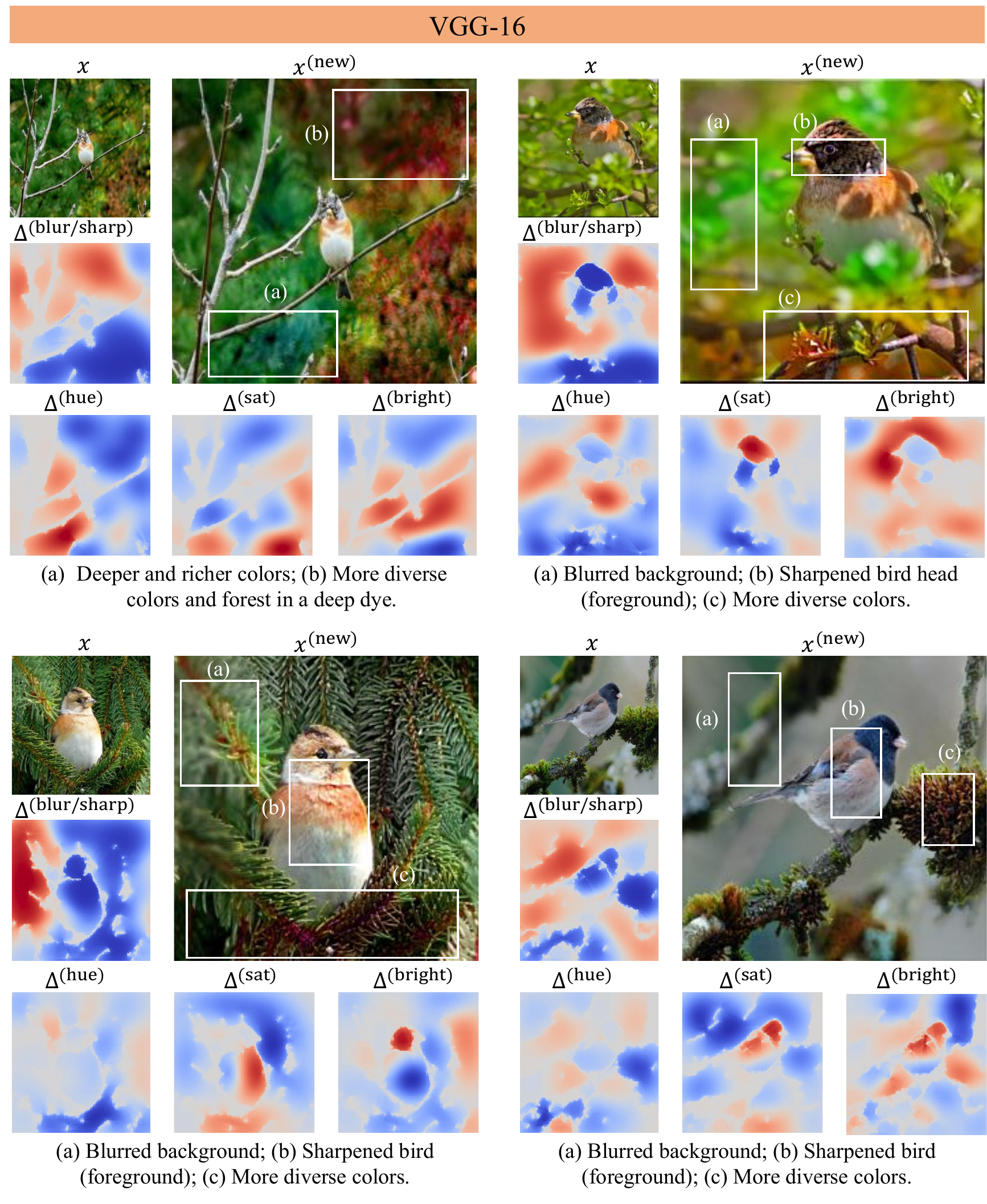}
	\caption{Images revised by using the VGG-16 model. The heatmap {\small$\Delta^{\text{(blur/sharp)}}$} denoted each pixel was blurred or sharpened, where the red color referred to blurring and the blue color corresponded to sharpening.
	Heatmaps {\small$\Delta^{\text{(hue)}}$}, {\small$\Delta^{\text{(sat)}}$}, and {\small$\Delta^{\text{(bright)}}$} indicated the increase (red) and the decrease (blue) of the hue/saturation/brightness revision, respectively.}
	\label{vgg_1}
\end{figure}

\begin{figure}[h!]
	\centering
	\includegraphics[width=\linewidth]{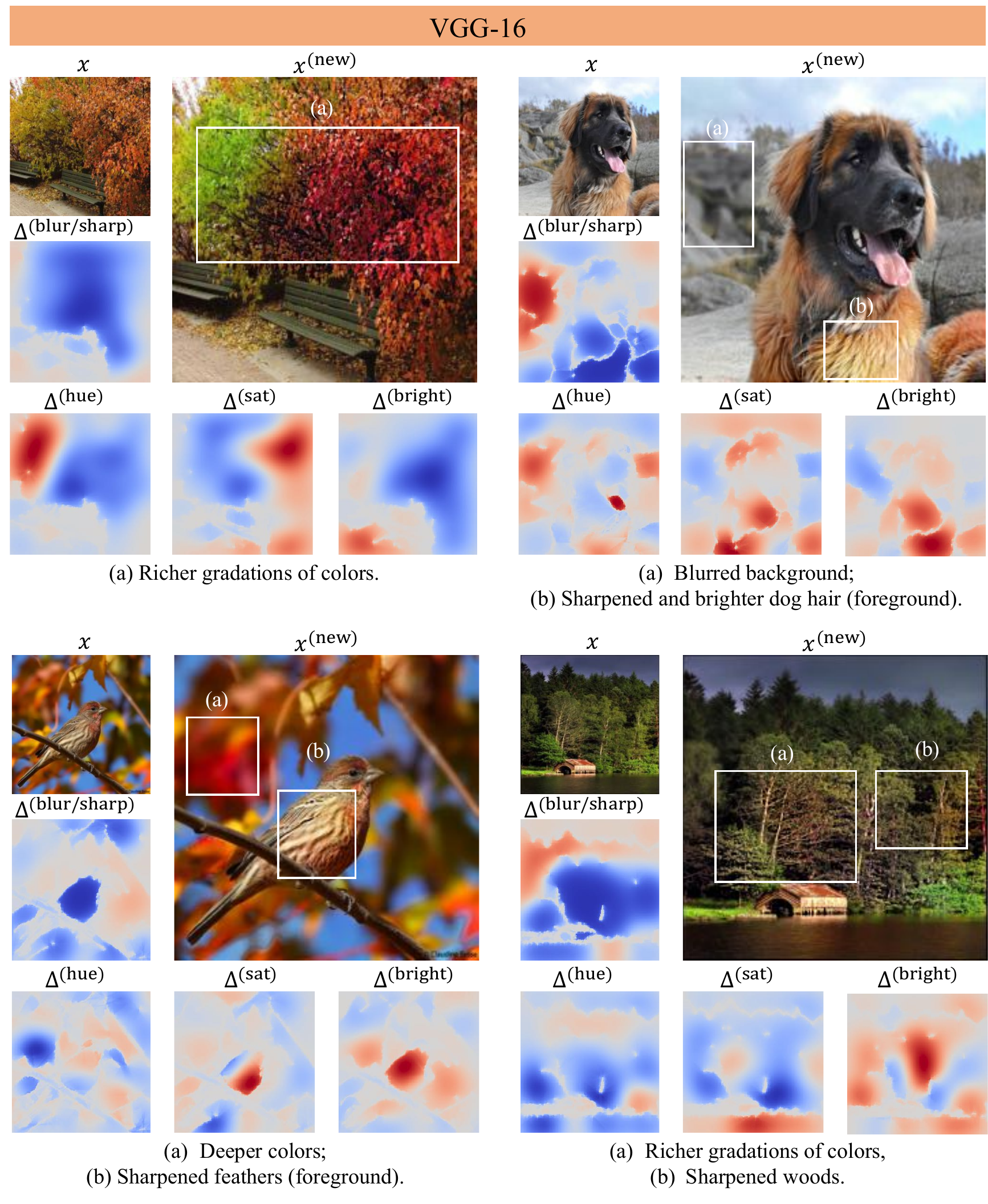}
	\caption{Images revised by using the VGG-16 model. The heatmap {\small$\Delta^{\text{(blur/sharp)}}$} denoted each pixel was blurred or sharpened, where the red color referred to blurring and the blue color corresponded to sharpening.
	Heatmaps {\small$\Delta^{\text{(hue)}}$}, {\small$\Delta^{\text{(sat)}}$}, and {\small$\Delta^{\text{(bright)}}$} indicated the increase (red) and the decrease (blue) of the hue/saturation/brightness revision, respectively.}
	\label{vgg_2}
\end{figure}

\begin{figure}[h!]
	\centering
	\includegraphics[width=\linewidth]{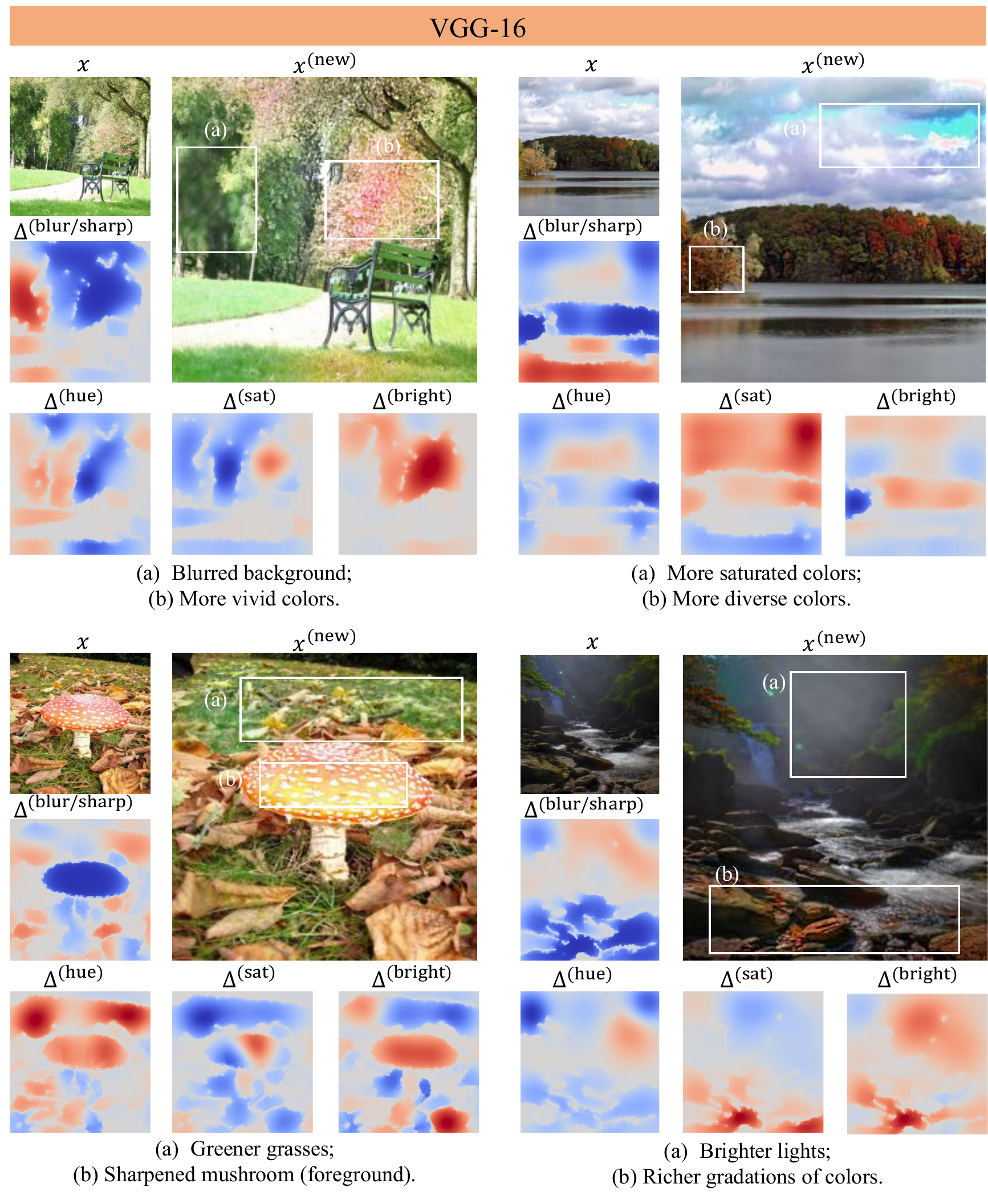}
	\caption{Images revised by using the VGG-16 model. The heatmap {\small$\Delta^{\text{(blur/sharp)}}$} denoted each pixel was blurred or sharpened, where the red color referred to blurring and the blue color corresponded to sharpening.
	Heatmaps {\small$\Delta^{\text{(hue)}}$}, {\small$\Delta^{\text{(sat)}}$}, and {\small$\Delta^{\text{(bright)}}$} indicated the increase (red) and the decrease (blue) of the hue/saturation/brightness revision, respectively.}
	\label{vgg_3}
\end{figure}

\begin{figure}[h!]
	\centering
	\includegraphics[width=\linewidth]{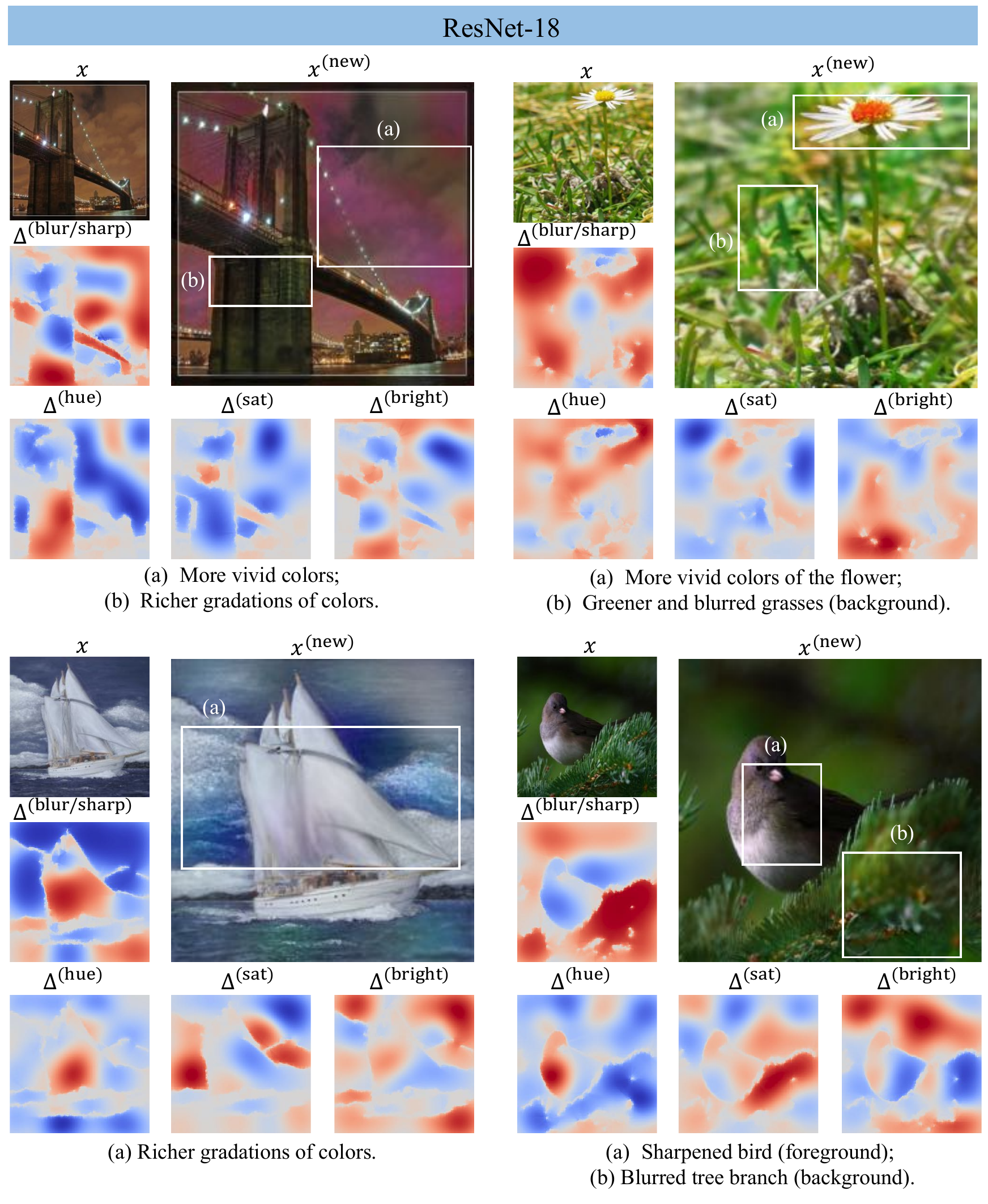}
	\caption{Images revised by using the ResNet-18 model. The heatmap {\small$\Delta^{\text{(blur/sharp)}}$} denoted each pixel was blurred or sharpened, where the red color referred to blurring and the blue color corresponded to sharpening.
	Heatmaps {\small$\Delta^{\text{(hue)}}$}, {\small$\Delta^{\text{(sat)}}$}, and {\small$\Delta^{\text{(bright)}}$} indicated the increase (red) and the decrease (blue) of the hue/saturation/brightness revision, respectively.}
	\label{res18_1}
\end{figure}

In this section, we present more results of the image revision.
As Fig.~\ref{vgg_1}, Fig.~\ref{vgg_2}, Fig.~\ref{vgg_3}, and Fig.~\ref{res18_1} show, the revised images were more aesthetic than the original ones from the following aspects. (1) \emph{The revised image usually had richer gradations of the color than the original image}. (2) \emph{The revised image often sharpened the foreground and blurred the background}.

\newpage

\newpage

\newpage

\newpage

\end{document}